\PassOptionsToPackage{table,dvipsnames}{xcolor}
\documentclass[10pt,twocolumn,letterpaper]{article}

\usepackage{iccv} 
\usepackage{multirow} 
\usepackage[table]{xcolor}
\usepackage{pifont}%
\newcommand{\cmark}{\ding{51}}%
\newcommand{\xmark}{\ding{53}}%

\usepackage{float}

\usepackage{float}
\usepackage{bbm}
\usepackage[table]{xcolor}
\usepackage{stmaryrd}

\definecolor{iccvblue}{rgb}{0.21,0.49,0.74}
\usepackage[pagebackref,breaklinks,colorlinks,allcolors=iccvblue]{hyperref}

\def\paperID{12610} 
\def\confName{ICCV}
\def\confYear{2025}

\title{\raisebox{-0.21\height}{\includegraphics[width=0.045\textwidth]{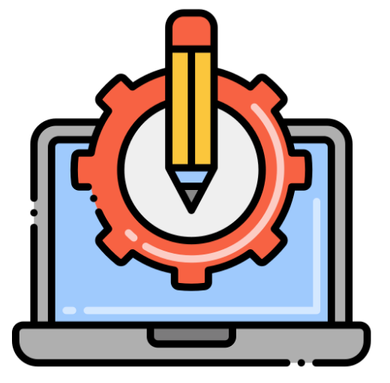}}~\texttt{CAD-Assistant}: Tool-Augmented VLLMs as Generic CAD Task Solvers}

\author{
    Dimitrios Mallis\textsuperscript{1} \\
    {\tt\small dimitrios.mallis@uni.lu} \and
    Ahmet Serda Karadeniz\textsuperscript{1} \\
    {\tt\small ahmet.karadeniz@uni.lu} \and
    Sebastian Cavada\textsuperscript{1} \\
    {\tt\small sebastian.cavada.dev@gmail.com} \and
    Danila Rukhovich\textsuperscript{1} \\
    {\tt\small danila.rukhovich@uni.lu} \and
    Niki Foteinopoulou\textsuperscript{1} \\
    {\tt\small niki.foteinopoulou@uni.lu} \and
    Kseniya Cherenkova\textsuperscript{1,2} \\
    {\tt\small kseniya.cherenkova@uni.lu} \and
    Anis Kacem\textsuperscript{1} \\
    {\tt\small anis.kacem@uni.lu} \and
    Djamila Aouada\textsuperscript{1} \\
    {\tt\small djamila.aouada@uni.lu} \and
    {\textsuperscript{1}SnT, University of Luxembourg \quad \textsuperscript{2}Artec3D, Luxembourg}
}

\begin{document}

\twocolumn[{%
\renewcommand\twocolumn[1][]{#1}%
\maketitle
\begin{center}
    \centering
    \vspace{-0.4cm}
    \includegraphics[width=\textwidth]{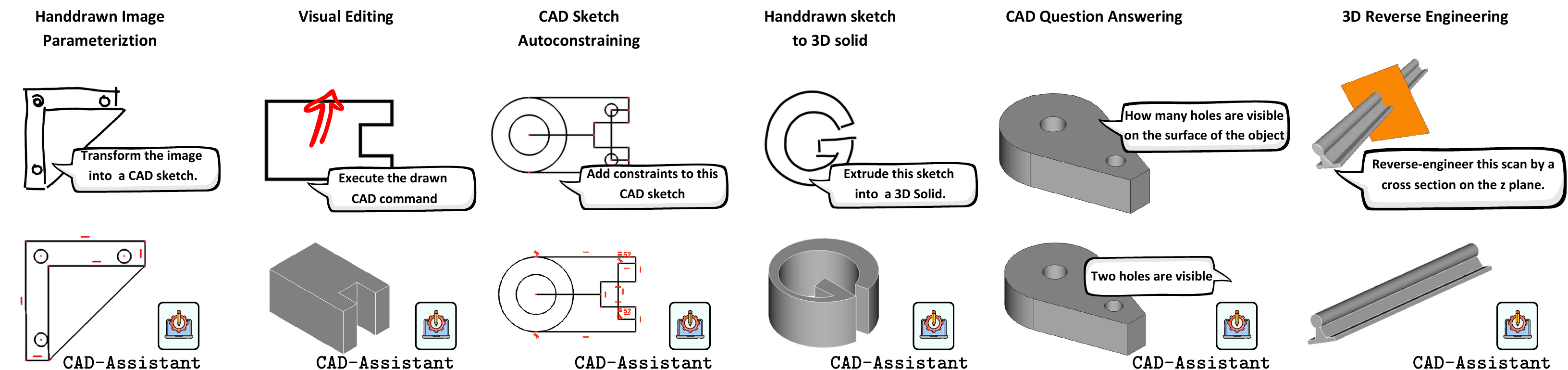}
    \captionof{figure}{\texttt{CAD-Assistant} is a tool-augmented VLLM framework for AI-assisted CAD. Our framework generates FreeCAD~\cite{FreeCAD} code that is executed within CAD software directly and can process multimodal inputs, including textual queries, sketches, drawn commands and 3D scans. This figure showcases various examples of generic CAD queries and the responses generated by \texttt{CAD-Assistant}. }
    \vspace{0.3cm}
    \label{teaserfig}
\end{center}%
}]

\maketitle

\begin{abstract}
We propose \texttt{CAD-Assistant}, a general-purpose CAD agent for AI-assisted design. Our approach is based on a powerful Vision and Large Language Model (VLLM) as a planner and a tool-augmentation paradigm using CAD-specific tools. \texttt{CAD-Assistant} addresses multimodal user queries by generating actions that are iteratively executed on a Python interpreter equipped with the FreeCAD~\cite{FreeCAD} software, accessed via its Python API. Our framework is able to assess the impact of generated CAD commands on geometry and adapts subsequent actions based on the evolving state of the CAD design. We consider a wide range of CAD-specific tools including a sketch image parameterizer~\cite{Karadeniz2024DAVINCIAS}, rendering modules, a 2D cross-section generator, and other specialized routines. \texttt{CAD-Assistant} is evaluated on multiple CAD benchmarks, where it outperforms VLLM baselines and supervised task-specific methods. Beyond existing benchmarks, we qualitatively demonstrate the potential of tool-augmented VLLMs as general-purpose CAD solvers across diverse workflows. Code implementation of the \texttt{CAD-Assistant} framework is publicly available~\small{\url{https://github.com/dimitrismallis/CAD-Assistant}}.
\end{abstract}    
\section{Introduction}

Computer-Aided Design (CAD) refers to the use of computer software to assist in the creation, modification, analysis, or optimization of a design~\cite{briere2012comparing}. Recently, there has been a significant research interest in the automation of CAD pipelines. Examples include, 3D reverse-engineering~\cite{mallis2023sharp, dupont2024transcad,khan2024cad,uy2022point2cyl}, CAD generation~\cite{seff2022vitruvion, xu2022deep,wu2021deepcad, Rukhovich2024CADRecodeRE}, edge parametrization~\cite{cherenkova2023sepicnet,zhu2023nerve}, CAD from multiview images~\cite{Hong2024MV2CylR3,you2024img2cad}, hand-drawn CAD sketch parametrization~\cite{Karadeniz2024PICASSOAF,Karadeniz2024DAVINCIAS} and text-guided CAD editing~\cite{Kodnongbua2023reparamCAD}. Still, most efforts to date have centered around fixed workflows, and the development of CAD agents to address generic tasks remains largely unexplored. In this work, we advocate that the creation of CAD agents capable of interacting with and supporting designers through the CAD process, would be a transformative advancement for the CAD industry.

As Vision and Large Language Models (VLLMs) continue to mature~\cite{Li2022BLIPBL,Achiam2023GPT4TR, Dai2023InstructBLIPTG, anthropic, Dubey2024TheL3, alayrac2022flamingo, liu2023llava, liu2023improvedllava}, they hold promise for enabling AI-assisted CAD design, particularly given that their very vast pre-training endow them with broad knowledge of design and manufacturing~\cite{Makatura2023HowCL}. Despite the identified potential, their ability to be used within computational design and manufacturing workflows remains severely constrained by weaknesses in geometric reasoning and handling of mathematical concepts~\cite{Hu2024VisualSS}. Indeed, VLLMs may struggle to semantically interpret the appearance of rendered objects from their corresponding CAD sequences~\cite{Qiu2024CanLL}. They may also fail to recognize spatial arrangements and the varied combinations of visual concepts~\cite{Sharma2024AVC} or correctly orient primitives and generate accurate placements~\cite{Makatura2023HowCL}. Their effectiveness in an agentic CAD setting is further hindered by the inherently unpredictable effects of CAD commands. High-level CAD operations, such as applying geometric constraints, fillet, chamfer, etc, can have complex and non-intuitive impacts on a model’s geometry and topology~\cite{seff2020sketchgraphs, seff2022vitruvion}, which is typically resolved by advanced CAD solvers. VLLMs cannot reliably predict the cumulative effects of the CAD commands they generate further limiting their practical usability in CAD workflows.

Recently, tool-augmentation has emerged as a prevailing strategy for addressing various shortcomings of foundational models and enhancing their performance in real-world applications~\cite{Hu2024VisualSS,Suris2023ViperGPTVI,lu2023chameleon,Shen2023HuggingGPTSA, Wu2023VisualCT}. Despite demonstrated effectiveness, VLLMs capable of composing and utilizing external tools have yet to be explored within the domain of CAD design. This work addresses this gap by introducing \texttt{CAD-Assistant}, a generic tool-augmented VLLM framework that integrates CAD-specific tools to effectively address the limitations of VLLMs in AI-assisted CAD. \texttt{CAD-Assistant} integrates a wide range of external CAD-specific modules, including a hand-drawn image parameterizer, rendering modules for multimodal CAD sequence understanding, a specialized utility for analysis of geometric constraints and a 2D cross-section generator for VLLM interaction with 3D scans.

Our framework leverages a VLLM-based planner and CAD-specific tool augmentation for generic CAD task solving. The planner generates CAD code actions, that are executed directly within the open-source CAD software FreeCAD~\cite{FreeCAD}, accessed via its Python API. Geometric reasoning is enhanced by dedicated CAD rendering and parameter serialization modules, enabling a more comprehensive multimodal representation of CAD models throughout the planning and reasoning process. Instead of solely relying on the effect prediction of complex CAD commands, our CAD agent inspects the evolving state of a design and refines or corrects actions based on the current CAD geometry. CAD-specific tools facilitate the processing of multimodal inputs, from text to hand-drawn sketches, precise CAD drawings, drawn commands and 3D scans.

\texttt{CAD-Assistant} is a \textit{training-free} framework that generates CAD code on an open-source CAD API, producing outputs that are both editable and highly interpretable. \texttt{CAD-Assistant} is also highly extensible and can operate across the diverse set of commands available in the FreeCAD API, requiring only a Python docstring to incorporate further capabilities. This is in contrast to the majority of CAD automation research focusing on the limited set of CAD operations captured in large-scale CAD datasets~\cite{wu2021deepcad, xu2022skexgen, li2024cad, xu2023hnc}. To address the lack of benchmarks for tool use akin to specialized sets commonly used in other domains~\cite{Lu2022LearnTE,Lu2022DynamicPL}, this work adopts an evaluation setting for generic CAD agents leveraging multiple existing CAD tasks. Evaluations are conducted for 2D and 3D CAD question answering, auto-constraining, and hand-drawn CAD sketch image parametrization. \texttt{CAD-Assistant} outperforms both VLLM baselines and supervised task-specific methods trained on large-scale datasets, despite being prompted in a zero-shot manner. Furthermore, we demonstrate the potential of \texttt{CAD-Assistant} beyond existing benchmarks by showcasing diverse use cases, including generating 3D solids from hand-drawn sketches, performing 3D reverse engineering from 3D scans via cross-section parameterization, and visual CAD design through semantically interpretable drawing commands (\eg sketching an extrusion operation). Example responses of the proposed \texttt{CAD-Assistant} on diverse multimodal queries are depicted in Figure~\ref{teaserfig}.

\vspace{0.1cm}
\noindent \textbf{Contributions}: The main contributions of this work can be summarized as follows:
\begin{enumerate}
    \item We introduce \texttt{CAD-Assistant}, the first tool-augmented VLLM framework for generic CAD task solving. Our framework is equipped with a diverse set of CAD-specific tools and can process multimodal inputs, including hand-drawn sketches and 3D scans.
    \item We demonstrate the effectiveness of tool-use for mitigating VLLMs' limitations on AI-assisted CAD. Geometic reasoning is enhanced by incorporating comprehensive multimodal representations of CAD models and enabling direct interaction with CAD software.
    \item We propose a highly extensible and training-free framework that can operate beyond the simple set of CAD commands captured on existing CAD datasets.
    \item We identify an evaluation setting for generic CAD agents based on existing benchmarks. The proposed zero-shot method outperforms baselines and task-specific approaches trained on large datasets. We also qualitatively demonstrate the potential of \texttt{CAD-Assistant} on a diverse set of real-world use cases. 
    
\end{enumerate}

\vspace{0.2cm}


\section{Related Work}
\label{sec:related_work}

\noindent \textbf{Foundation Models for CAD}: Recently, there has been increasing research interest in the use of foundation models on CAD-related applications. CAD-Talk~\cite{Yuan2023CADTalkAA} introduces a framework for semantic CAD code captioning using multi-view photorealistic renderings of CAD models along with part-segmentation, powered by foundation models~\cite{Kirillov2023SegmentA, Caron2021EmergingPI}. Taking a similar path, QueryCAD~\cite{kienle2024querycad} proposes an open-vocabulary CAD part segmentation from images leveraging segmentation foundation models and LLMs to perform CAD-related question-answering for robotic applications. CADLLM~\cite{cadllm} proposes a T5 model~\cite{Raffel2019ExploringTL} finetuned on the SketchGraphs~\cite{seff2020sketchgraphs} dataset of 2D CAD sketches for sketch auto-completion. CadVLM~\cite{Wu2024CadVLMBL} extends CADLLM~\cite{cadllm} to the visual domain, incorporating a visual modality for CAD sketch auto-completion, autoconstraining and image-guided generation. CADReparam~\cite{Kodnongbua2023reparamCAD} uses VLLMs to infer meaningful variation spaces for parametric CAD models, re-parameterizing them to enable exploration along design-relevant axes. Img2CAD~\cite{you2024img2cad} utilizes a VLLM to reverse engineer objects from images, predicting the specific CAD command types needed to model each part of the object accurately. Badagabettu~\etal~\cite{Badagabettu2024Query2CADGC}, focus on text-guided generation of CAD models as CADQuery code, while LLM4CAD~\cite{li2024llm4cad} use a similar approach to generate 3D CAD models from text and image inputs. Related to ours is the training-free method of ~\cite{alrashedygenerating} focusing on CAD model generation. Authors introduce a verification process to ensure the validity of generated models, but do not explore tool augmentation. Our investigation diverges from these task-specific approaches as it shifts the focus on tool-augmentation for mitigating the limitation of VLLMs on AI-assisted CAD. \texttt{CAD-Assistant} is the first \textit{general-purpose} framework for CAD design, able to process multimodal prompts and address diverse CAD use cases.

\vspace{0.1cm}
\noindent \textbf{Tool-augmented VLLMs}: Recently there has been growing interest in enhancing LLMs and VLLM performance via augmentation with external tools~\cite{Zeng2022SocraticMC, Gupta2022VisualPC, Suris2023ViperGPTVI, lu2023chameleon,Shen2023HuggingGPTSA, Wu2023VisualCT, Yang2023MMREACTPC, Hu2024VisualSS}. The field is further propelled by the emergence of benchmarks, namely ScienceQA~\cite{Lu2022LearnTE} and TabMWP~\cite{Lu2022DynamicPL}, which are well-suited for evaluating the effectiveness of tool-use. 
Tool-use offers several benefits~\cite{Qin2023ToolLW}, such as reducing hallucinated knowledge~\cite{Shuster2021RetrievalAR}, providing real-time information~\cite{lu2023chameleon}, enhancing domain expertise~\cite{Nakano2021WebGPTBQ} and producing interpretable outputs by making intermediate steps explicit~\cite{Gupta2022VisualPC, Suris2023ViperGPTVI}. Planning is commonly performed via instructions in natural language~\cite{Gupta2022VisualPC, lu2023chameleon} or Python code generation~\cite{Suris2023ViperGPTVI, Hu2024VisualSS}, and tool set might include search engines~\cite{Nakano2021WebGPTBQ, Komeili2021InternetAugmentedDG, lu2023chameleon}, calculators~\cite{Cobbe2021TrainingVT, Parisi2022TALMTA}, external APIs~\cite{Patil2023GorillaLL}, vision modules~\cite{Gupta2022VisualPC,Suris2023ViperGPTVI}, Hugging Face models~\cite{Shen2023HuggingGPTSA}, Azure models~\cite{Yang2023MMREACTPC} or LLM created tools~\cite{Cai2023LargeLM}. Despite the vast potential of tool-augmented LLMs and VLLMs for CAD-related applications, the space remains unexplored. To our knowledge, this work is the first investigation on tool-augmented VLLMs for AI-assisted CAD.

\vspace{0.1cm}
\noindent \textbf{VLLMs as Geometrical Reasoners}: 
In order to advance tool-augmented VLLMs for AI-assisted CAD, it is crucial for the VLLMs planner to semantically recognize and precisely identify and manipulate individual elements within parametric geometries. This type of precision is an essential skill when interfacing with CAD software. Naturally, this raises the question: \textit{Can large vision language models understand symbolic graphics programs?} In that direction, Yi~\etal~\cite{Yi2018NeuralSymbolicVD} explored incorporating symbolic structure as prior knowledge for enhancing visual question answering. More recently, Sharma~\etal~\cite{Sharma2024AVC} examined visual program generation and recognition, showing that while shape generation often relies on memorizing prototypes from training data, shape recognition demands a deeper understanding of primitives. Qi~\etal~\cite{Qiu2024CanLL} introduced SGPBench, a question-answering benchmark designed to assess the semantic understanding and consistency of symbolic graphics programs, including CAD models. This benchmark evaluates the extent of LLMs' ability to semantically comprehend and reason about geometric structures. 
While \cite{Qiu2024CanLL} applied instruction tuning to improve visual program understanding, our work emphasizes general-purpose VLLMs, demonstrating that factors like serialization and parametrization strategies for formatting geometry and multimodal representation of a CAD model can significantly expand VLLMs' capacity for geometric reasoning.

\section{The proposed \texttt{CAD-ASSISTANT}}
\label{sec:method}

\subsection{General Framework}

This section provides an overview of \texttt{CAD-ASSISTANT}. Our framework comprises the following three components:

\vspace{0.1cm}
\noindent
\textbf{Planner}: The planner $\mathcal{P}$ is modelled by a VLLM capable of advanced reasoning. Following~\cite{Hu2024VisualSS}, on each timestep $t$, the planner analyses the current context $c_t$ and generates a plan $p_t$ and an action $a_t$ that implements $p_t$. In this work, we employ GPT-4o~\cite{Achiam2023GPT4TR} as the core framework planner.

\vspace{0.1cm}
\noindent
\textbf{Environment}: 
We utilize the Python interpreter as the primary environment $\mathcal{E}$ for executing the generated action $a_t$ at time $t$. Additionally, $\mathcal{E}$ integrates CAD software~\cite{FreeCAD} as a foundational component for AI-assisted CAD applications. On each timestep, $t$, the environment provides feedback $e_t$ of the current state of the CAD design.

\begin{figure}[t]
\setlength{\belowcaptionskip}{-0.2cm}
    \centering
\includegraphics[width=0.9\linewidth]{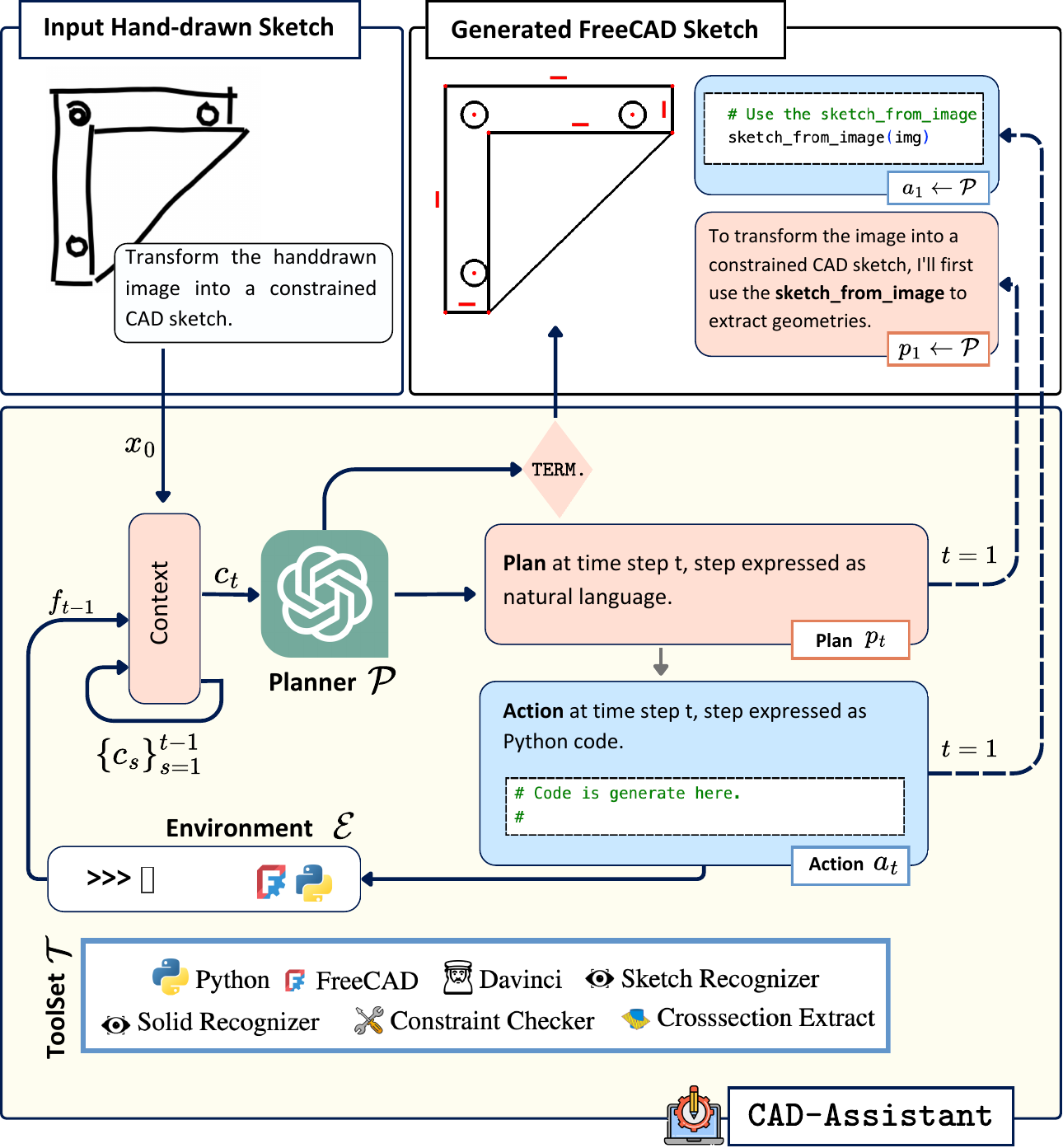}
    \vspace{-0.3cm}
    \caption{Overview of \texttt{CAD-Assistant} framework. A multimodal user request is provided as context to a VLLM planner $\mathcal{P}$. At step $t$, the planner generates a plan $p_t$ and an action $a_t$ (\textit{python code}). The action is executed on an environment $\mathcal{E}$ and the generated execution output $f_t$ is fed back to the planner, enabling generation for the next timestep.}
    \label{fig:overview}
\end{figure}

\begin{figure}[t]
\setlength{\belowcaptionskip}{-0.6cm}
    \centering
\includegraphics[width=0.9\linewidth]{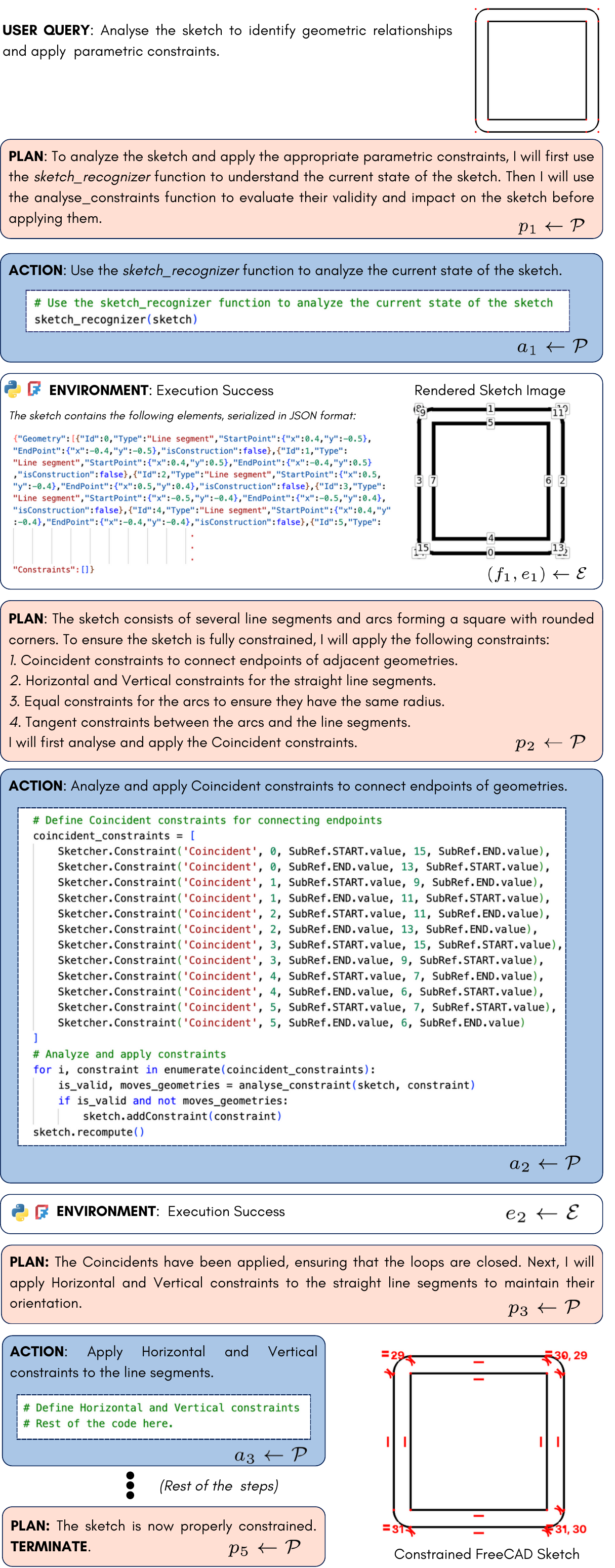}
    \vspace{-0.2cm}
    \caption{Execution flow for autoconstraining. The sketch recognizer function is utilized for multimodal CAD understanding. Constraints are generated over multiple timesteps.}
    \label{fig:autoconstraining}
\end{figure}

\vspace{0.1cm}
\noindent
\textbf{Tool Set}: \texttt{CAD-ASSISTANT} utilizes a set $\mathcal{T}~=~{\{\mathcal{T}_i\}}_{i=1}^N$ of $N$ CAD-specific tools, suitable for AI-Assisted CAD. These include
standard Python libraries, modules of the FreeCAD Python API~\cite{FreeCAD} to interface CAD commands, and other useful CAD-specific tools and Python routines. \texttt{CAD-ASSISTANT} can be formalized as follows: Given a multimodal $x_0$ input user query, on each timestep $t$, the planner $\mathcal{P}$ generates:

\vspace{-0.2cm}
\begin{equation}
    p_t \leftarrow \mathcal{P}(x_0; c_{t-1}, \mathcal{T}) \ ,
\end{equation}
\begin{equation}
    a_t \leftarrow \mathcal{P}(p_t; c_{t-1}, x_0, \mathcal{T}) \ ,
\end{equation}

\noindent
where $p_t$ is the current plan in natural language, and $a_t$ is the current action formulated as Python code. Then, the generated action $a_t$ is executed on the framework's environment: 

\begin{equation}
    (f_{t}, e_t) \leftarrow \mathcal{E}(a_t; e_{t-1}, \mathcal{T}, x_0) \ ,
\end{equation}

\noindent
where $f_t$ is the output of the code execution, and $e_t$ is the new state of the CAD design. Note that $f_t$ can include both textual and visual outputs of the execution, \eg list of CAD geometries in \texttt{.json} format or the rendering of the current state of the CAD object. Finally, the context is updated as:

\vspace{-0.2cm}
\begin{equation}
    c_{t+1} \leftarrow \texttt{concat}(f_t, \{c_s\}_{s=1}^{t}) \ ,
    \label{eq:eq_concat}
\end{equation}

\begin{table}[t]
\setlength{\belowcaptionskip}{-0.5cm}
    \centering
        \setlength{\tabcolsep}{3pt}
    \small
    \begin{tabular}{l|p{0.27\textwidth}}
        \toprule
        \textbf{Module Type} & \textbf{Module Description} \\
        \midrule
        \raisebox{-0.2\height}{\includegraphics[width=0.025\textwidth]{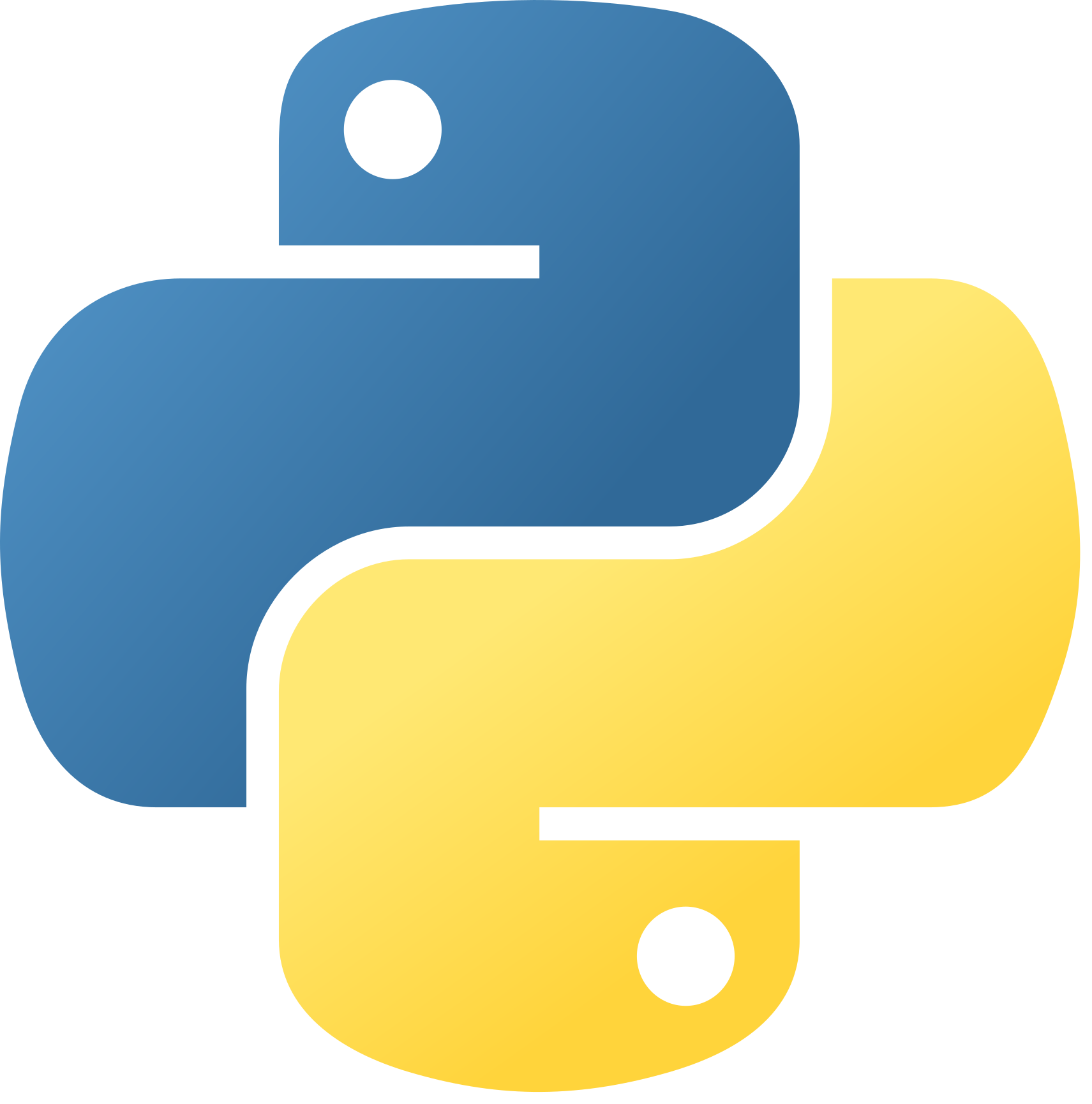}} Python & Action format and logical operations.\\
        \midrule
        \raisebox{-0.2\height}{\includegraphics[width=0.02\textwidth]{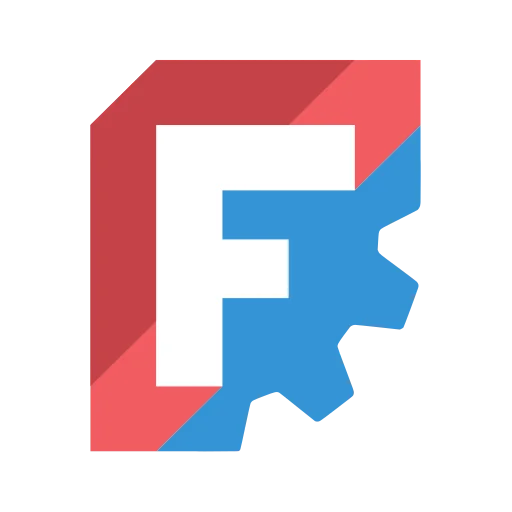}} FreeCAD & Integration with CAD software~\cite{FreeCAD}.\\
        \midrule
        \raisebox{-0.2\height}{\includegraphics[width=0.02\textwidth]{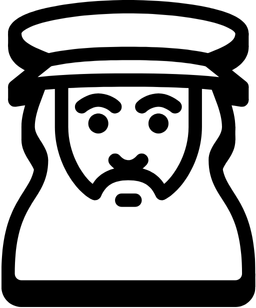}} Sketch Parameterizer & Hand-drawn sketch image to CAD sketch based via ~\cite{Karadeniz2024DAVINCIAS}.\\
        \midrule
        \raisebox{-0.05\height}{\includegraphics[width=0.02\textwidth]{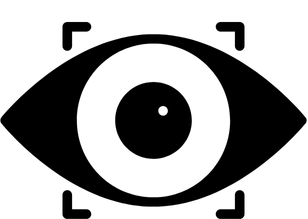}} Sketch Recognizer & Renders sketch and plots parameters.\\
        \midrule
        \raisebox{-0.2\height}{\includegraphics[width=0.02\textwidth]{assets/recognizerlogo.png}} Solid Recognizer & Renders a 3D CAD model and plots parameters.\\
        \midrule
        \raisebox{-0.2\height}{\includegraphics[width=0.02\textwidth]{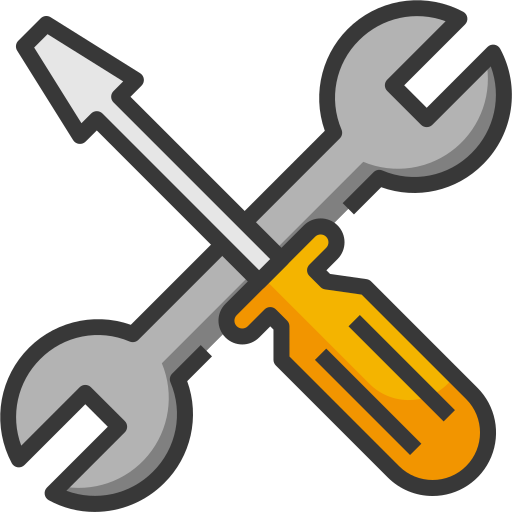}} Constraint Checker & Analyzes geometric constraints.\\
        \midrule
        \raisebox{-0.2\height}{\includegraphics[width=0.02\textwidth]{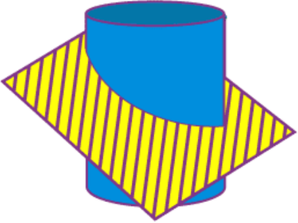}} Crosssection Extract & Generates an image of a cross section from a 3D mesh.\\
        \bottomrule
    \end{tabular}
    \vspace{-0.2cm}
    \caption{Overview of CAD-specific tools.}
\label{tab:modulesummary}
\end{table}

\noindent
concatenating the previous context with the current code execution output and is supplied to $\mathcal{P}$ for plan generation of timestep $t+1$. This process iterates for an arbitrary $T$ number of steps until the planner $\mathcal{P}$ concludes that the request $x_0$ has been successfully addressed. At that point, $\mathcal{P}$ generates $p_{T}$, a special \texttt{TERMINATE} plan that indicates the completion of \texttt{CAD-ASSISTANT}'s response. An illustration of the proposed execution flow is provided in Figure~\ref{fig:overview} and an example of the agents' trajectory as a response to an autoconstraining request is provided in Figure~\ref{fig:autoconstraining}.


\definecolor{light-gray}{gray}{0.95}
\newcommand{\code}[1]{\colorbox{light-gray}{\texttt{#1}}}

\subsection{CAD-specific Tool-set}

\texttt{CAD-ASSISTANT} includes a set of $N$ CAD-specific tools or modules. Each tool is defined by its method signature and the docstring~\cite{Hsieh2023ToolDE} that disambiguates its use. Modules $\mathcal{T}_i$ are instantiated via their Python interface with arguments generated by $\mathcal{P}$ as part of the action $a_t$. Notably, actions are formulated as Python code, as in~\cite{Suris2023ViperGPTVI, Hu2024VisualSS}, rather than the natural-language instructions advocated by recent works~\cite{lu2023chameleon, Gupta2022VisualPC}. This design choice allows for direct use of the FreeCAD API. Moreover, the generated action $a_t$ can access the parameters of the CAD models' state $e_t$ and perform logical and computational operations, which is highly advantageous for design tasks (see also section \ref{sec:verification} of supplementary). Our CAD-specific tool set is summarized in Table \ref{tab:modulesummary} and a detailed overview of each tool is provided in supplementary.

\section{Experiments}
\label{ref:evaluations}

This section outlines the experiments conducted to validate the effectiveness of \texttt{CAD-Assistant}. 


\subsection{Strategies for Effective Geometric Reasoning}
\label{ref:ablations}

Effective geometric reasoning is an essential requirement for the development of generic CAD agents. However, VLLMs have shown limited ability to geometrically comprehend and mathematically reason about CAD programs~\cite{Qiu2024CanLL,Sharma2024AVC}. Previous work has explored symbolic instruction tuning~\cite{Qiu2024CanLL} for addressing this limitation. In contrast, we shift our focus on tool augmentation as a training-free alternative to enhance geometric reasoning. This subsection examines CAD representations that can be derived using external tools and improve VLLMs' understanding of CAD programs. Specifically, we study the following factors of a CAD representation:

\noindent
\textbf{Parametrization Strategy:} Parametric geometries can be represented by different sets of parameters. For instance, a line could use start / end points or an angle and length relative to a reference. We compare the implicit parametrization approach of~\cite{Qiu2024CanLL} to the point-based primitive representation of~\cite{Karadeniz2024PICASSOAF}. We also explore over-parametrization, where a redundant set of parameters is used per geometry. More details about this comparison are provided on supplementary.

\begin{table}[t]
\setlength{\belowcaptionskip}{-0.5cm}
    \centering
    \setlength{\tabcolsep}{8pt}
    \begin{tabular}{llc}
        \multicolumn{3}{c}{\textbf{2D CAD SGPBench - Sketch in Textual Format}}\\
        \addlinespace[2pt]
        \toprule
         \textbf{Serialization} & \textbf{Parametarization}  & \textit{Acc} \\
        \midrule
        \\[-10pt]
        \multicolumn{3}{l}{\underline{\textit{SGPBench~\cite{Qiu2024CanLL} format}}}\\
         \\[-10pt]
        Serialized Graph & Implicit
        &  0.674 \\ 
        \\[-5pt]
        \multicolumn{3}{l}{\underline{\textit{Standardized CAD Sketch formats}}}\\
        \\[-10pt]
        DXF~\cite{dxf} &   & 0.671 \\      
        OCA~\cite{oca} & & 0.707 \\    
        \\[-5pt]
        \multicolumn{3}{l}{\underline{\textit{Serialization Strategy (Tabular formats)}}}\\
        \\[-10pt]
          CSV & Point-based 
        &  0.703 \\       
         Markdown & Point-based 
        &  0.706 \\        
        HTML & Point-based 
        &  0.710 \\      
        \\[-5pt]
        \multicolumn{3}{l}{\underline{\textit{Serialization Strategy (Schema-embedded formats)}}}\\
        \\[-10pt]
        Serialized Graph & Point-based 
        &  0.744 \\ 
        JSON & Point-based 
        &  0.748 \\      
        \\[-5pt]
        \multicolumn{3}{l}{\underline{\textit{Parametarization Strategy}}}\\
        \\[-10pt]
        JSON & Point-based
        &  \textbf{0.748} \\        
        JSON & Overparametarized
        &  0.747 \\    
    \bottomrule
\addlinespace[8pt]
        \multicolumn{3}{c}{\textbf{2D CAD SGPBench - Sketch as a Rendering}}\\
                \addlinespace[2pt]
        \toprule
        \multicolumn{2}{l}{\textbf{CAD Sketch Image Type}}  & \textit{Acc} \\
        \midrule
        \multicolumn{2}{l}{Hand-drawn Sketch} & 0.616 \\   
         \multicolumn{2}{l}{Precise Rendering} & \textbf{0.754} \\
    \bottomrule
    \end{tabular}
    \vspace{-0.2cm}
    \caption{Investigation of prompting strategies on geometric reasoning. We report performance for GPT-4o in terms of accuracy on the 2D partition of SGPBench~\cite{Qiu2024CanLL}. \textbf{(Top)} Impact of Parametrization and serialization on CQA performance. \textbf{(Bottom)} Performance from hand-drawn and precise rendering of a CAD sketch.}
\label{tab:ablation_sgp}
\end{table}

\noindent
\textbf{Serialization Strategy:} The serialization format used to convert the parametric geometry into text can impact the planner's ability to understand the geometry. Motivated by recent work on text-based serialization methods for tabular data~\cite{Fang2024LargeLM}, we compare commonly used formats such as \textit{CSV}, \textit{Markdown}, \textit{HTML}, and \textit{JSON}.

\noindent
\textbf{Rendering-based Reasoning:} We investigate visual representations for geometric reasoning by providing the VLLM planner with 2D renderings of the CAD sketch or 3D solid.

To examine the impact of the above strategies on CAD program understanding and geometric reasoning, we experiment on the CAD question answering benchmark SGPBench~\cite{Qiu2024CanLL}. This benchmark comprises multiple-choice questions and captures three types of graphical programs, \ie,  SVG, CAD sketches, and 3D CAD models. For this experiment, we report accuracy on the 2D CAD subset. This subset is derived from $700$ CAD sketches from SketchGraphs~\cite{seff2020sketchgraphs}. A VLLM planner (GPT-4o) is provided with a textual description of a 2D CAD sketch and tasked with answering a multiple-choice question about the design.

In Table~\ref{tab:ablation_sgp}~(\textbf{top}), we analyze the effect on the performance of the parametrization and serialization strategies used to parse the CAD sketch into a textual format. Firstly, we observe that schema-embedded representation like \textit{JSON} performs better than tabular formats. Note that this is in contrast with recent work~\cite{Sui2023TableML}, where \textit{HTML} was identified as the optimal serialization for tabular data. Secondly, GPT-4o demonstrates high sensitivity to geometry parametrization. The implicit parametrization used in SGPBench~\cite{Qiu2024CanLL} significantly under-performs compared to a point-based parametrization for geometric primitives as in~\cite{Karadeniz2024PICASSOAF}. Overall, using a \textit{JSON} serialization along with the point-based parametrization from~\cite{Karadeniz2024PICASSOAF} leads to substantial improvements over the original SGPBench format and other text-based CAD sketch formats, such as \textit{DXF} and \textit{OCA}. While over-parameterizing the sketches results in a negligible drop in performance \wrt a point-based parameterization, we argue that it is safer to opt for over-parameterization as other tasks might benefit from it. Furthermore, as shown in Table~\ref{tab:ablation_sgp}~(\textbf{bottom}), rendering-based question answering surpasses the performance reported for text-based recognition. Following these findings, we equip the \texttt{CAD-Assistant} with a specialized recognition tools that generate an over-parameterized JSON representation of CAD models as well as renderings of 2D CAD sketch or 3D solid for comprehensive multimodal geometric reasoning.

\begin{table}[t]
\setlength{\belowcaptionskip}{-0.5cm}
    \centering
    \setlength{\tabcolsep}{6pt}
    \begin{tabular}{llcc}
        \toprule
         \textbf{Method} & \textbf{Planner} & \textit{2D Acc} &  \textit{3D Acc} \\
         \midrule
         \multirow{3}{*}{SGPBench~\cite{Qiu2024CanLL}} & GPT-4 mini & 0.594& 0.737 \\
          & GPT-4 Turbo & 0.674& 0.762\\
          & GPT-4o & 0.686 & 0.782\\
        
         \midrule
         \multirow{3}{*}{\texttt{CAD-ASSISTANT}} & GPT-4 mini & 0.614& 0.783\\
          & GPT-4 Turbo & 0.741 & 0.825\\
          & GPT-4o & \textbf{0.791} & \textbf{0.857}\\
        \bottomrule
    \end{tabular}
    \vspace{-0.2cm}
    \caption{Comparison for the proposed \texttt{CAD-ASSISTANT} to baselines for CQA on the 2D and 3D subsets of SGPBench~\cite{Qiu2024CanLL}. For \texttt{CAD-Assistant} performance is reported for different planners.}
\label{tab:results_qa}
\end{table}

\subsection{CAD Benchmarks and Experimental Setup}
\label{sec:benchmarks}

As a generic framework, \texttt{CAD-Assistant} can be conditioned to perform a wide range of tasks related to CAD design. Given the lack of specialized evaluation benchmarks for CAD agents, this work adapts an evaluation setting based on the following existing CAD tasks.


\vspace{0.1cm}
\noindent
\textbf{CAD Question Answering}: As in subsection~\ref{ref:ablations}, quantitative evaluations of CAD Question Answering (CQA) is performed on the recently introduced SGPBench~\cite{Qiu2024CanLL}.  We do not provide the CAD code as part of the prompt as in~\cite{Qiu2024CanLL}. Instead, the CAD sketch or model is pre-loaded into a FreeCAD project file, allowing \texttt{CAD-Assistant} to utilize the FreeCAD integration and CAD-specific tools to understand the design and answer questions. This experimental setup simulates a real-world question-answering environment where a CAD designer can ask open-ended questions about the design to support the iterative design process. 
We report accuracy on the 2D and 3D CAD sets.

\vspace{0.1cm}
\noindent
\textbf{Autoconstraining}: Parametric constraints are a key component of feature-based CAD modeling~\cite{mallis2023sharp} and a widely adapted mechanism for explicit capturing of design intent~\cite{Zhang2009DesignII, Otey2018RevisitingTD}. Given a CAD sketch of $n$ parametric primitives ~\hbox{$\{\mathbf{p}_1, \mathbf{p}_2, ..., \mathbf{p}_n\} \in \mathcal{P}^n$} (lines, arcs, circles, points)
the goal of autoconstraining is to infer a set of parametric constraints $\{\mathbf{c}_i\}_{i=1}^m \in \mathcal{C}^m$ applied on these primitives. Each constraint $\mathbf{c}_i$ is composed of constraint type, participating primitives $\mathbf{p_i}$, $\mathbf{p_j}$ and subreferences $(s_{i}, s_{j})$ specifying the point of application (\eg start, end, center). In contrast to the evaluation setting of~\cite{seff2020sketchgraphs, seff2022vitruvion}, we incorporate the application of the geometric solver (CAD software) to determine the final configuration of sketch primitives. Performance is measured in terms of Primitive F1 Score \textit{(PF1)} and Constraint F1 Score \textit{(CF1)} as in~\cite{yang2022discovering}. \textit{PF1} defines a true positive as a primitive with the correct type and parameters within five quantization units, and for \textit{CF1} a constraint is considered a true positive only if all associated primitives are also correctly predicted. Quantitative evaluations are performed on SketchGraphs~\cite{seff2020sketchgraphs}. We use the test set of~\cite{seff2022vitruvion} and evaluate on a subset of $700$ CAD sketches due to the resource intensive nature GPT4-o API requests.


\vspace{0.1cm}
\noindent
\textbf{Hand-drawn CAD sketch Parameterization}: Given a binary sketch image $\mathbf{X} \in \{0,1\} ^{h \times w}$, sketch parameterization aims to recover the complete constrained CAD sketch $(\{\mathbf{p}_i\}_{i=1}^n,\{\mathbf{c}_i\}_{i=1}^m)$. We report parametric accuracy computed on quantized primitive tokens as in~\cite{seff2022vitruvion,Karadeniz2024PICASSOAF} after solving the CAD sketch. We also compute bidirectional Chamfer Distance (CD) on the image space. 
 For evaluation, we use the same test split as in the autoconstraining task. For hand-drawn sketch synthesis, we follow the strategy of~\cite{seff2022vitruvion}.


\vspace{0.1cm}

\subsection{Experimental Results}

We evaluate the performance of \texttt{CAD-Assistant} on the benchmarks described in the previous section.

\noindent
\textbf{CAD Question Answering:} \texttt{CAD-Assistant} is able to interact directly with a CAD model via its integration with CAD software and is tasked with answering a question about the design. 
Results for CQA on SGPBench~\cite{Qiu2024CanLL} are reported in Table \ref{tab:results_qa}. For this experiment, we also report the performance of the GPT-4 mini and GPT-4 Turbo models as planners. We observe that by leveraging available tools such as the Python interpreter and the comprehensive multimodal representation of CAD models generated via the recognizer tools, \texttt{CAD-Assistant} improves CQA performance for both CAD sketches and 3D CAD models, thus highlighting the potential of tool-use for CAD understanding. Notably, for the smaller GPT-4 mini, the performance gain from \texttt{CAD-Assistant} is marginally above (2D subset) or on-par (3D subset), emphasizing the need for pairing tool-augmented frameworks with a powerful VLLM.

\begin{table}[t]
\setlength{\belowcaptionskip}{-0.2cm}
    \centering
    \setlength{\tabcolsep}{8pt}
    \begin{tabular}{llcc}
        \toprule
         \textbf{Method} & \textbf{Type} &\textit{PF1}~$\uparrow$ & \textit{CF1}~$\uparrow$ \\
         \midrule
         GPT-4o & \textit{zero-shot}& 0.693 & 0.274\\
         Vitruvion~\cite{seff2022vitruvion} & \textit{supervised} & 0.706 & 0.238\\

         \texttt{CAD-ASSISTANT} & \textit{zero-shot}& \textbf{0.979} & \textbf{0.484}\\ 
        \bottomrule
    \end{tabular}
    \vspace{-0.2cm}
    \caption{Evaluation on the task of autoconstraining. Performance is measured in terms of \textit{PF1} and \textit{CF1} on the SketchGraphs~\cite{seff2020sketchgraphs}.}
\label{tab:results_autoconstraining}
\end{table}

\begin{table}[t]
\setlength{\belowcaptionskip}{-0.6cm}
    \centering
    \setlength{\tabcolsep}{3pt}
    \begin{tabular}{cccccc}
        \toprule
        \multicolumn{2}{c}{\textbf{CAD-Specific Tools}} & \multicolumn{2}{c}{\textbf{Prompting}} &\textit{PF1}~$\uparrow$ & \textit{CF1}~$\uparrow$\\
         \textit{MMrecog} & \textit{ConstrCheck} & \textit{Demonstr} & \textit{Docstr} & & \\
\cmidrule(lr){1-2} 
\cmidrule(lr){3-4} 
         \xmark & \xmark & \textit{0-shot} & \cmark & 0.726 & 0.318\\
        \cmark & \xmark & \textit{0-shot} & \cmark & 0.747 & 0.329\\ 
          \cmark & \cmark & \textit{0-shot} & \cmark & 0.979 & 0.484\\ 
          \cmark & \cmark & \textit{5-shot} & \xmark & 0.981 & 0.514\\ 
         \cmark & \cmark & \textit{5-shot} & \cmark & \textbf{0.984} & \textbf{0.529}\\ 
        \bottomrule
    \end{tabular}
    \vspace{-0.2cm}
    \caption{Impact of CAD-specific tools and prompting strategies for \texttt{CAD-Assistant} on the autoconstraining task.}
\label{tab:ablation_autoconstraining}
\end{table}

\vspace{0.1cm}
\noindent
\textbf{Autoconstraining:} We evaluate our method on the task of CAD sketch autoconstraining~\cite{seff2022vitruvion}. \texttt{CAD-Assistant} is prompted to apply a set of parametric constraints with proper design intent to a CAD sketch preloaded into a FreeCAD project file, similar to the CQA setup. Performance is compared to a GPT-4o baseline and the constraint generation model Vitruvion~\cite{seff2022vitruvion}, trained on a large-scale dataset~\cite{seff2020sketchgraphs}. Results are reported on Table \ref{tab:results_autoconstraining}. Note that the autoconstraining performance is reported after solving the predicted constraints with a CAD solver. As we are operating within CAD software, the CAD solver enforces the predicted constraints (\eg, orthogonality between two lines) on CAD sketches, adjusting the parameters of the affected primitives accordingly (\eg, modifying the parameters of the two lines). We observe that both the baseline and~\cite{seff2022vitruvion} tend to generate poorly parameterized constraints, which may lead to the arbitrary repositioning of primitives when applied by the CAD solver, as evidenced by the low \textit{PF1} values. In contrast, \texttt{CAD-Assistant} effectively utilizes tools to interact with the CAD software, assesses the impact of constraints, and preserve the integrity of the geometry. Notably, constraints generated by \texttt{CAD-Assistant} result in a high \textit{CF1} score despite \textit{zero-shot} prompting, further underscores the broad understanding of \texttt{CAD-Assistant} in CAD design. In Table \ref{tab:ablation_autoconstraining}, we investigate the impact of tools relevant to auto-constraining on the effectiveness of \texttt{CAD-Assistant}. We find that both the multimodal sketch recognizer (\textit{MMrecog}) and the constraint checker module (\textit{ConstrCheck}) contribute to performance gains. Table \ref{tab:ablation_autoconstraining} also compares prompting strategies for the proposed framework. While we primary focus on zero-shot prompting, which promotes agentic behavior by eliminating the need for CAD designers to create tailored examples for unique use cases, we find that a few high quality demonstrations can further enhance performance as shown by the results for \textit{5-shot} prompting. 


\begin{table}[t]
\setlength{\belowcaptionskip}{-0.2cm}
    \centering
    \setlength{\tabcolsep}{9pt}
    \begin{tabular}{lcc}
        \toprule
         \textbf{Method} & \textit{Acc}~$\uparrow$  & \textit{CD}~$\downarrow$  \\
         \midrule
         Vitruvion~\cite{seff2022vitruvion}  & 0.659 & 1.586\\
        Davinci~\cite{Karadeniz2024DAVINCIAS}  & \textbf{0.789} &1.184\\
        \texttt{CAD-ASSISTANT}  & 0.784 & \textbf{0.680}\\ 
        \bottomrule
    \end{tabular}
    \vspace{-0.2cm}
    \caption{Evaluation on the task of hand-drawn image parametrization. Comparison against the task-specific models of~\cite{seff2022vitruvion, Karadeniz2024DAVINCIAS}. }
\label{tab:results_parameterization}
\end{table}

\begin{figure}[t]
\setlength{\belowcaptionskip}{-0.6cm}
    \centering
\includegraphics[width=0.8\linewidth]{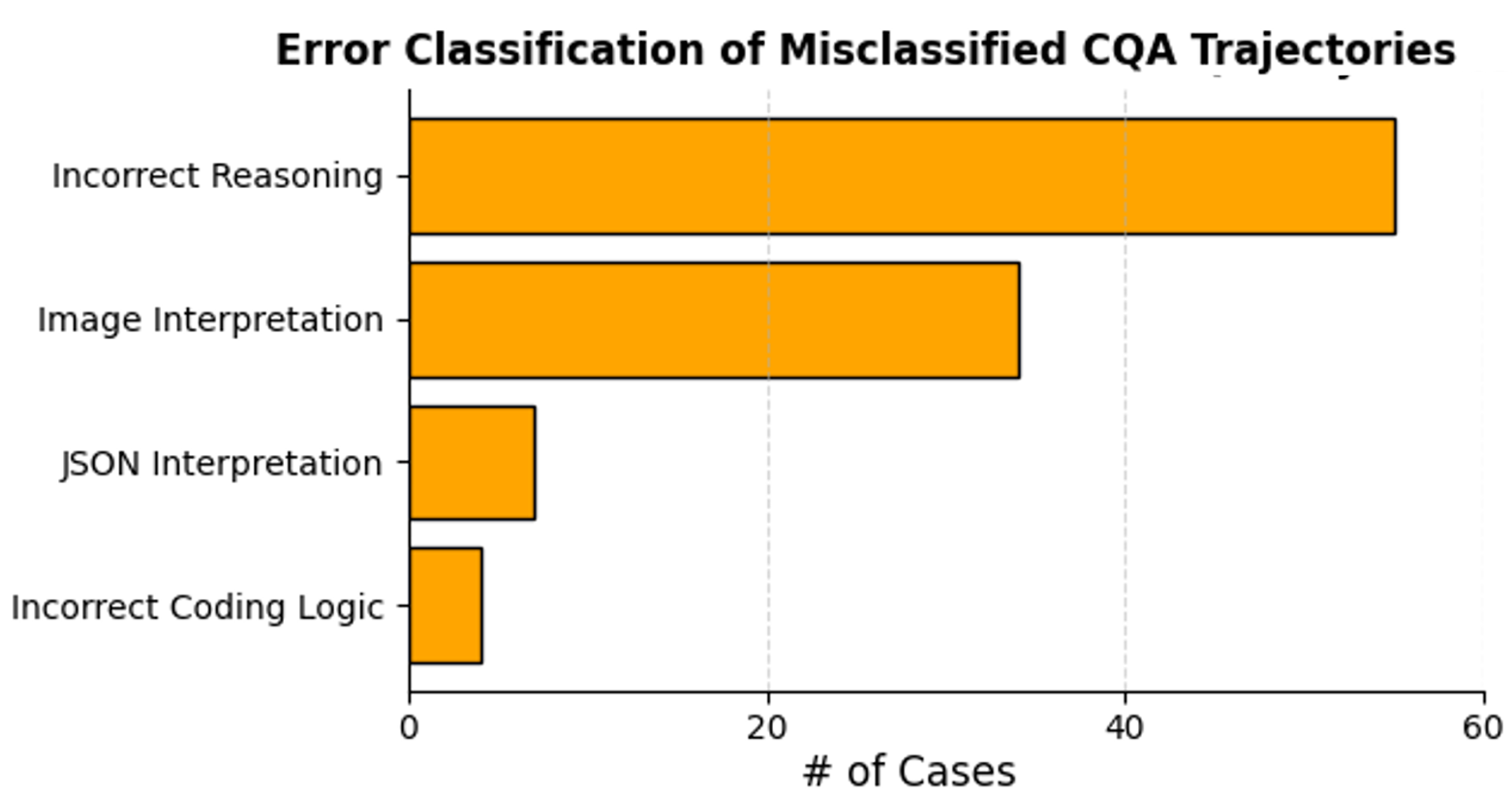}
\vspace{-0.2cm}
\caption{Classification of failure case types for erroneous responses in the CAD Question Answering task.}
\label{fig:failure}
\end{figure}

\vspace{0.1cm}
\noindent
\textbf{Hand-drawn CAD sketch Parameterization:} Our framework utilizes the sketch parameterization tool that processes hand-drawn inputs to generate a textual description of primitives and constraints, as well as the constraint analysis module to assess the impact of constraints on CAD geometry. Performance is compared to task-specific models in Table \ref{tab:results_parameterization}. We observe that \texttt{CAD-Assistant} effectively translates the text-based parameterization recovered by the sketch parameterizer (based on~\cite{Karadeniz2024DAVINCIAS}) into a FreeCAD sketch, resulting in high accuracy. Additionally, it successfully applies constraints without compromising the solved geometry, as evidenced by the reduction in CD.

\begin{figure*}
\setlength{\belowcaptionskip}{-0.6cm}
    \centering
    \includegraphics[width=0.93\textwidth]{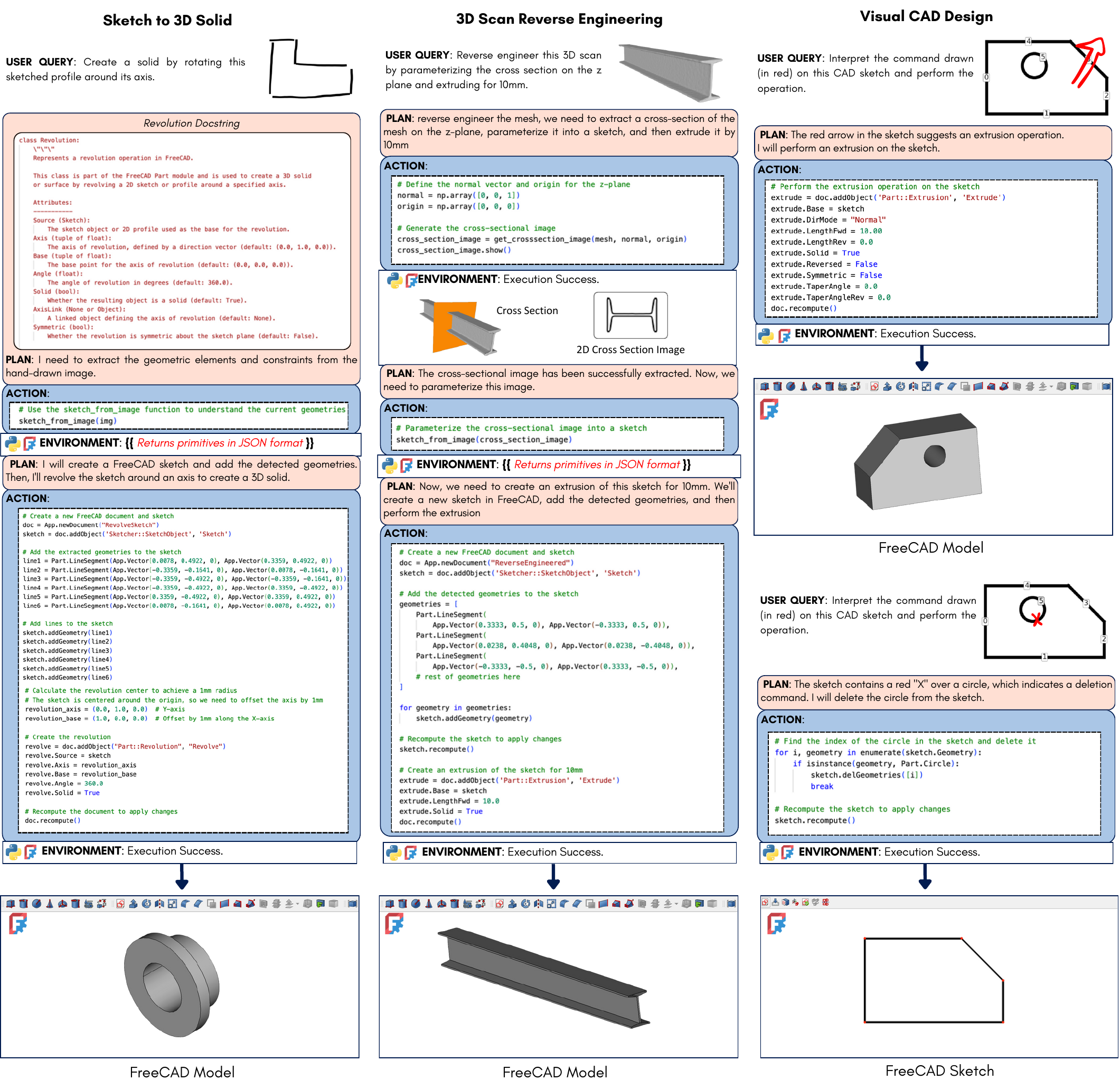}
    \vspace{-0.3cm}
    \caption{Real-world CAD use cases. (\textbf{Left}) The \texttt{CAD-Assistant} generated a 3D solid conditioned on a handdrawn sketch image. (\textbf{Center}) Our method reconstructs a 3D scan via cross-section parameterization. (\textbf{Right}) The \texttt{CAD-Assistant} semantically interprets the drawn operation and fulfills user requests directly without composing CAD-specific tools.}
    \label{fig:beyond}
\end{figure*}

\vspace{0.1cm}
\noindent
\textbf{Human Evaluations.} We conduct a failure case analysis on the CQA task, shown in Figure~\ref{fig:failure}. A human annotator reviews 100 agent trajectories associated with misclassified answers and categorizes the error types. Most failures are caused from incorrect reasoning by the VLLM or misinterpretation of tool-generated renderings, such as visually confusing a trapezoid with a triangle. To evaluate the effectiveness of \texttt{CAD-Assistant} in tool usage, two human annotators examine 200 autoconstraining / parameterization trajectories for tool-use validity. The analysis shows a high validity rate of 98.5\%, with the few errors observed primarily due to incorrect use of the FreeCAD API.

\subsection{Exploring New Capabilities}
\label{sec:beyond}


\noindent
\textbf{Beyond Simplified CAD Commands:} Research on common CAD tasks generally focuses on the limited sets of CAD commands captured by large-scale datasets~\cite{seff2020sketchgraphs, xu2022deep}. As a train-free framework, \texttt{CAD-Assistant} can leverage the full range of commands available within the FreeCAD API requiring only the corresponding docstring. On Figure~\ref{fig:beyond}~\textit{(left)} and Figure~\ref{fig:beyond_v2}~\textit{(supplementary)} we showcase examples of our method utilizing the CAD commands \textit{Revolution} and \textit{Fillet} that are not included in existing datasets~\cite{xu2022deep}.

\noindent
\textbf{Real-world use cases:} Tool augmentation allows interaction with multimodal inputs such as sketches and 3D scans. Figure~\ref{fig:beyond} (\textit{center}) showcases \texttt{CAD-Assistant}'s ability to process 3D scans along with textual queries to extract cross-sections, parameterize features, and reverse engineer CAD models from scans. In Figure~\ref{fig:beyond} (\textit{right}), the VLLM planner determines to semantically interpret drawn operation directly without utilizing additional CAD-specific tools for fulfilling user requests. Note that generated FreeCAD code is interpretable, editable and easily extendable.

\section{Conclusion}
\label{ref:conclusions}
In this work, we propose \texttt{CAD-Assistant}, a generic tool-augmented CAD agent using CAD-specific tools. Our framework responds to multimodal queries via generated actions that are executed in a python interpreter integrated with FreeCAD. We assess \texttt{CAD-Assistant} on diverse CAD benchmarks and demonstrate the potential of tool-augmented VLLMs in real-world CAD workflows.

\section{Acknowledgements}
The present work is supported by the National Research Fund (FNR), Luxembourg, under the BRIDGES2021/IS/16849599/FREE-3D project and by Artec3D.

{
    \small
    \bibliographystyle{ieeenat_fullname}
    \bibliography{main}
}

\clearpage
\setcounter{page}{1}
\maketitlesupplementary

This supplementary material includes various details that were not reported in the main paper due to space constraints. To demonstrate the benefit of the proposed \texttt{CAD-Assistant}, we also expand our qualitative evaluation.

\section{CAD-specific Tool-set}
This section provides a detailed discussion of the CAD-specific tool set utilised by the proposed framework. \texttt{CAD-ASSISTANT} is equiped with the following tools:

\vspace{0.1cm}
\noindent
\textbf{Hand-drawn Image Parameterizer}: To enable visual sketching, we employ a task-specific model for hand-drawn image parameterization~\cite{Karadeniz2024DAVINCIAS}. This module extracts parameters and constraints as text, allowing \texttt{CAD-Assistant} to reuse primitive parameters for CAD code generation.

\vspace{0.1cm}
\noindent
\textbf{CAD Sketch Recognizer}: We equip \texttt{CAD-Assistant} with a CAD sketch recognition utility. This routine returns both a summary of geometries and parametric constraints in \texttt{.json} format, along with a visual rendering of the CAD sketch. The rendered sketch image includes numeric markers of the primitive ID overlayed on the rendered geometries. Motivated by~\cite{yang2023setofmark}, this approach enhances visual grounding for GPT-4o,~\ie its ability to associate visual content with the textual description of primitives.

\vspace{0.1cm} \noindent \textbf{3D Solid Recognizer}: 
For CAD model recognition, we also incorporate a 3D solid recognizer that generates a \texttt{.json} summary of model parameters (for both sketch and extrusion operations) along with visual renderings of the 3D solid from four different angles, providing a multimodal representation of structure and geometry.

\vspace{0.1cm}
\noindent
\textbf{Constraint Checker}: We include a dedicated function that evaluates the parameters of a parametric constraint to determine its validity and whether it causes movement in geometric elements. The constraint analyzer facilitates effective interaction with the CAD solver by assessing the impact of commands like parametric constraints on geometry.

\vspace{0.1cm}
\noindent \textbf{Cross-section Extract}: Cross-sections are critical components of CAD reverse engineering workflows~\cite{cross_sections}. \texttt{CAD-Assistant} includes a specialized routine for  2D cross-section images from 3D scans across 2D planes.

\vspace{0.1cm}
\noindent
\textbf{FreeCAD API}:  \texttt{CAD-Assistant} is integrated with the open-source FreeCAD software~\cite{FreeCAD} via the FreeCAD Python API. This API enables programmatic control over the majority of commands available to designers and access to the current state of the CAD design. In this work, we consider a range of components from the Sketcher and Part modules of the FreeCAD API, focusing on CAD sketching, the addition and manipulation of primitives, geometric constraints, and extrusion operations for constructing 3D solids. A summary of the exact classes, methods and class attributes of the FreeCAD API integrated with \texttt{CAD-Assistant} is provided in the supplementary.%

\vspace{0.1cm}
\noindent
\textbf{Python}: Beyond facilitating actions $a_t$, the planner can utilize Python as a tool to conduct essential logical and mathematical operations, such as calculating segment lengths, determining angles, and deriving parameter values.

\section{System Details}

\texttt{CAD-Assistant}'s implementation is based on the Autogen~\cite{Wu2023AutoGenEN} programming framework for Agentic AI. We report \texttt{CAD-Assistant}'s performance with \small{\texttt{gpt-4o-mini-2024-07-18}}, \small{\texttt{gpt-4-turbo-2024-04-09}} and \small{\texttt{gpt-4o-2024-08-06}} as VLLM planners, accessed via API calls.

\section{CAD Representations}
\label{sec:sketches}
In this section, we provide a formally introduction of 2D CAD sketches and 3D CAD models.

\subsection{Constrained CAD Sketches}
A constraint CAD sketch is commonly represented by a graph $\mathcal{G} =(\mathcal{P}^n, \mathcal{C}^m)$ comprising a set of $n$ primitive nodes~\hbox{$\{\mathbf{p}_1, \mathbf{p}_2, ..., \mathbf{p}_n\} \in \mathcal{P}^n$} and $m$ edges between nodes $\{\mathbf{c}_1, \mathbf{c}_2, ..., \mathbf{c}_m\} \in \mathcal{C}^m$ denoting geometric constraints. Primitives $\mathbf{p}_i$ are of type line $\mathbf{l}_i$, arc $\mathbf{a}_i$, circle $\mathbf{c}_i$ or points $\mathbf{d}_i$. VLLM and LLM planners can be sensitive to the parameterization strategy followed for representing $\mathbf{p}_i$. 
This work conducts an investigation on the impact of sketch parameterization on visual program understanding in black-box VLLMs presented in section \ref{ref:ablations} where we compare the following parameterization strategies:

\vspace{0.1cm}
\noindent
\textbf{Implicit}: This is the parameterization strategy utilized for representation of 2D CAD sketches by the SGPBench~\cite{Qiu2024CanLL}. Primitives $p_i$ are represented as follows:

\begin{table}[H]
    \centering
    \setlength{\tabcolsep}{20pt}
    \resizebox{\linewidth}{!}{
    \begin{tabular}{lr}
        \toprule
         $\mathbf{a}_i = (x_{c}, y_{c}, v_{x}, v_{y}, b_{wc}, \theta_s, \theta_e) \in \mathbb{R}^4 \times \{0, 1\} \times [0, 2\pi)^2$& \\
        $\mathbf{c}_i = (x_{c}, y_{c},r) \in \mathbb{R}^3$& \\
        $\mathbf{l}_i = (x_{p}, y_{p},v_{x}, v_{y}, d_s, d_e) \in \mathbb{R}^6$& \\
        $\mathbf{d}_i= (x_p, y_p) \in \mathbb{R}^2$& \\
        \bottomrule
    \end{tabular}
    }

    \caption{Implicit parameterization strategy for arcs $\mathbf{a}_i$, circles $\mathbf{c}_i$, lines $\mathbf{l}_i$ and points $\mathbf{p}_i$.}
\label{tab:types}
\end{table}


where and $(x_c, y_c)$ denotes center point coordinates, $(d_s, d_e)$ are signed start/end point distances to a point $(x_p, y_p)$, the unit direction vector is denoted as $(v_{x}, v_{y})$, radius is denoted with $r$, $(\theta_s, \theta_e)$ are the start/end angles to the unit direction vector in radians and $b_{wc}$ is a binary flag indicating if the arc is clockwise.

\vspace{0.1cm}
\noindent
\textbf{Point-based}: We contrast the implicit parameterization to the point-based approach from~\cite{seff2022vitruvion, Karadeniz2024PICASSOAF, Karadeniz2024DAVINCIAS} as described on the following table. 

\begin{table}[H]
    \centering
    \setlength{\tabcolsep}{40pt}
    \resizebox{\linewidth}{!}{
    \begin{tabular}{lr}
        \toprule
        $\mathbf{a}_i = (x_{s}, y_{s},x_{m}, y_{m},x_{e}, y_{e}) \in \mathbb{R}^6$ & \\
        $\mathbf{c}_i = (x_{c}, y_{c},r) \in \mathbb{R}^3$& \\
        $\mathbf{l}_i = (x_{s}, y_{s},x_{e}, y_{e}) \in \mathbb{R}^4$& \\
        $\mathbf{d}_i = (x_p, y_p) \in \mathbb{R}^2$& \\
        \bottomrule
    \end{tabular}
    }

    \caption{Point-based parameterization strategy for arcs $\mathbf{a}_i$, circles $\mathbf{c}_i$, lines $\mathbf{l}_i$ and points $\mathbf{p}_i$.}
\label{tab:types}
\end{table}

where $(x_s, y_s)$, $(x_m, y_m)$, $(x_e, y_e)$ are start, middle and end point coordinates and $r$ is the radius.

\vspace{0.1cm}
\noindent
\textbf{Overparameterized}: This strategy is a simple combination of the implicit and point-based parameterization.

\begin{table}[H]
    \centering
    \setlength{\tabcolsep}{1pt}
    \resizebox{\linewidth}{!}{
    \begin{tabular}{l}
        \toprule
        $\mathbf{a}_i = (x_{c}, y_{c}, v_{x}, v_{y}, x_{s}, y_{s}, x_{m}, y_{m}, x_{e}, y_{e}, b_{wc}, \theta_s, \theta_e) 
\in \mathbb{R}^{10} \times \{0, 1\} \times [0, 2\pi)^2$\\
     $\mathbf{c}_i = (x_{c}, y_{c},r) \in \mathbb{R}^3$\\
     $\mathbf{l}_i = (x_{p}, y_{p},v_{x}, v_{y}, d_s, d_e, x_{s}, y_{s},x_{e}, y_{e}) \in \mathbb{R}^{10}$\\
     $\mathbf{d}_i = (x_p, y_p) \in \mathbb{R}^2$\\
        \bottomrule
    \end{tabular}
    }

    \caption{Overparameterized parameterization strategy for arcs $\mathbf{a}_i$, circles $\mathbf{c}_i$, lines $\mathbf{l}_i$ and points $\mathbf{p}_i$.}
\label{tab:types}
\end{table}

We identify the overparameterized strategy as the safest approach, as it enables the VLLM planner to leverage a broader and more diverse set of parameters, better accommodating the varying requirements of different input queries. In addition to parametric primitives $\mathbf{p}_i$, a CAD sketch incorporates constraints defined by CAD designers, ensuring that future modifications propagate coherently throughout the design. A constraint is defined as an undirected between primitives $\mathbf{p}_i$ and $\mathbf{p}_j$. They might also include subreferences $(s_{i}, s_{j}) \in \llbracket 1..4 \rrbracket^2$, to specify whether the constraint is applied on \textit{start}, \textit{end}, \textit{middle} point, or \textit{entire} primitive for both $\mathbf{p}_i$ and $\mathbf{p}_j$. Note that some constraints may involve only a single primitive $\mathbf{p}_i$ (\eg a vertical line); in such cases, the constraint is defined as the edge between the primitive and itself. In this work we consider the following types of constraints: \textit{coincident, parallel, equal, vertical, horizontal,  perpendicular, tangent}.

\subsection{CAD Models}
Following the feature-based CAD modeling paradigm~\cite{mallis2023sharp, xu2022deep}, a CAD model $\mathbf C \in \mathcal{C}$ is constructed as a sequence of design steps. In this work, evaluation is performed on CAD models from the 3D partition of SGPBench~\cite{Qiu2024CanLL} sourced from the DeepCAD dataset~\cite{xu2022deep}. These models are constructed exclusively via a \textit{sketch-extrude} strategy, where 2D CAD sketches $\mathcal{G}_i$ are followed by extrusion operations that turns the sketch into a 3D volume. Extrusions include the following parameters:

\begin{table}[H]
    \centering
    \setlength{\tabcolsep}{10pt}
    \resizebox{0.8\linewidth}{!}{
    \begin{tabular}{lr}
        \toprule
        \textit{Parameter Description} & \textit{Parameter Notation}\\
        \midrule
        Sketch plane orientation &$(\theta, \phi, \gamma) \in [0, \pi] \times [0, 2\pi)^2$\\
        Sketch plane translation & $(\tau_x, \tau_y, \tau_z) \in \mathbb{R}^3$\\
        sketch scale & $\sigma \in \mathbb{R}$\\
        Extrusion distance & $(d_{-}, d_{+})\in \mathbb{R}^2$\\
        Extrusion type & $\beta \in \{1,2,3,4\}$\\
        \bottomrule
    \end{tabular}
    }

    \caption{Extrusion Parameters description.}
\label{tab:types}
\end{table}

\noindent where extrusion type $\beta$ can be among \textit{new}, \textit{cut}, \textit{join} and \textit{intersect}.

\subsection{Parameter Quantization}
\label{sec:quantization}
Unlike prior task-specific models for CAD-related tasks such as hand-drawn sketch parameterization~\cite{seff2022vitruvion, Karadeniz2024PICASSOAF, Karadeniz2024DAVINCIAS}, CAD sketch generation~\cite{seff2022vitruvion}, or 3D CAD model generation~\cite{xu2022deep}, the \texttt{CAD-Assistant} does not rely on the common practice of parameter quantization. Typically, these methods use a $6$-bit uniform quantization scheme to convert continuous sketch and extrusion parameters into discrete tokens, enabling prediction through transformer-based sequence architectures trained with cross-entropy loss~\cite{seff2022vitruvion, Karadeniz2024PICASSOAF, Karadeniz2024DAVINCIAS,xu2022deep}. In contrast, the \texttt{CAD-Assistant} employs a VLLM planner that directly regresses primitive and extrusion parameters as continuous numerical values. We apply the 6-bit uniform quantization to the outputs of \texttt{CAD-Assistant}, to facilitate direct comparisons with task-specific methods for autoconstraining and hand-drawn sketch parameterization reported on section \ref{sec:benchmarks} of the main paper.

\begin{figure}[h]
    \centering
\includegraphics[width=\linewidth]{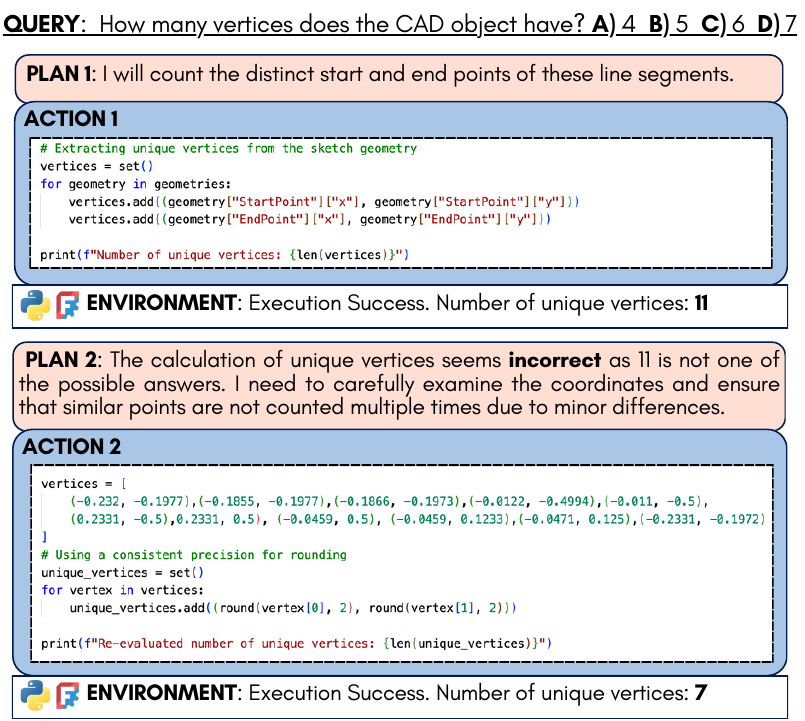}
    \caption{Correction of an inaccurate answer for a CQA example.}
    \label{fig:correction}
\end{figure}

\section{Verification of Responses}
\label{sec:verification}
The proposed workflow allows for verifying incorrect responses. The generated plan can be updated based on intermediate code execution results, including error logs (see $f_t$ in Eq.~\ref{eq:eq_concat}). Figure~\ref{fig:correction} shows that when an error occurs (\ie, answer 11 is not among the possible choices of the question), the planner detects this mistake and updates the plan accordingly. Exploring the potential of combining tool-augmentation with more advanced planning and verification algorithms (\eg ~\cite{alrashedygenerating}) is left as interesting future work.

\section{Costs}
The proposed \texttt{CAD-Assistant} utilizes a GPT-4o planner accessed through API calls. Table \ref{tab:costs} provides a summary of the costs associated with each user query across CAD benchmarks.

\begin{figure}[h]
    \centering
\includegraphics[width=.8\linewidth]{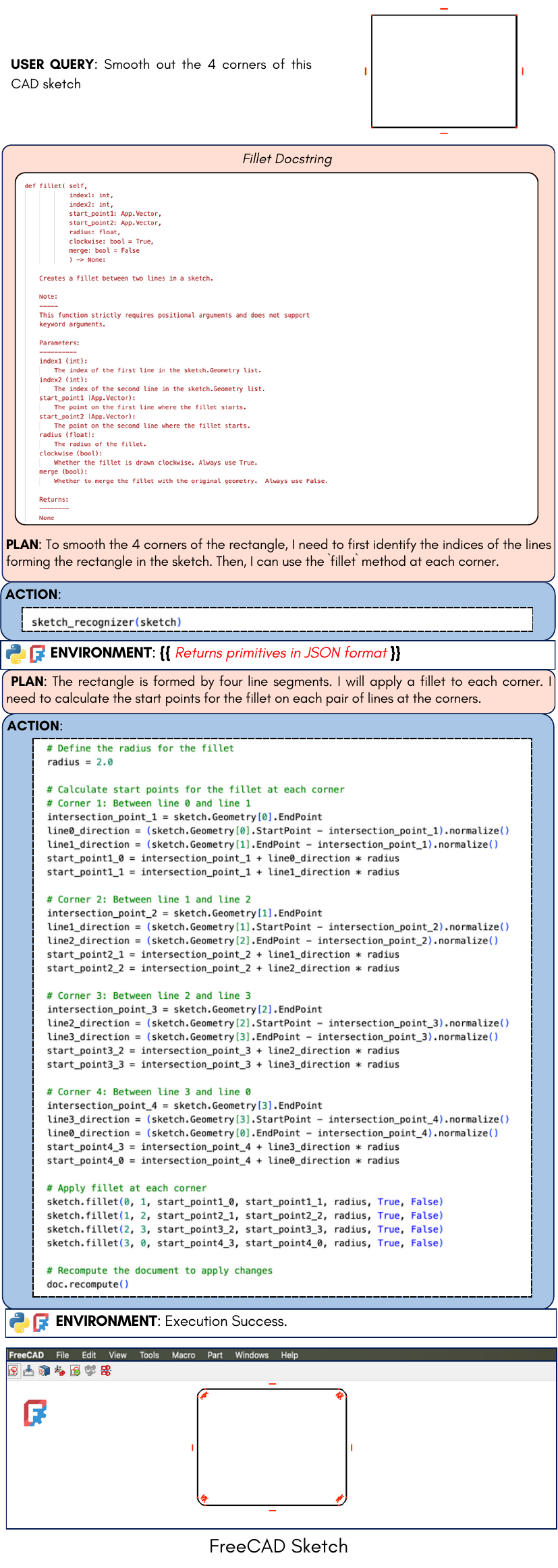}
    \caption{Example of the proposed \texttt{CAD-Assistant} utilizing the \textit{Fillet} CAD command.}
    \label{fig:beyond_v2}
\end{figure}

\section{\texttt{CAD-Assistant} Prompts}

\begin{figure*}[h]
    \centering
\includegraphics[width=\linewidth]{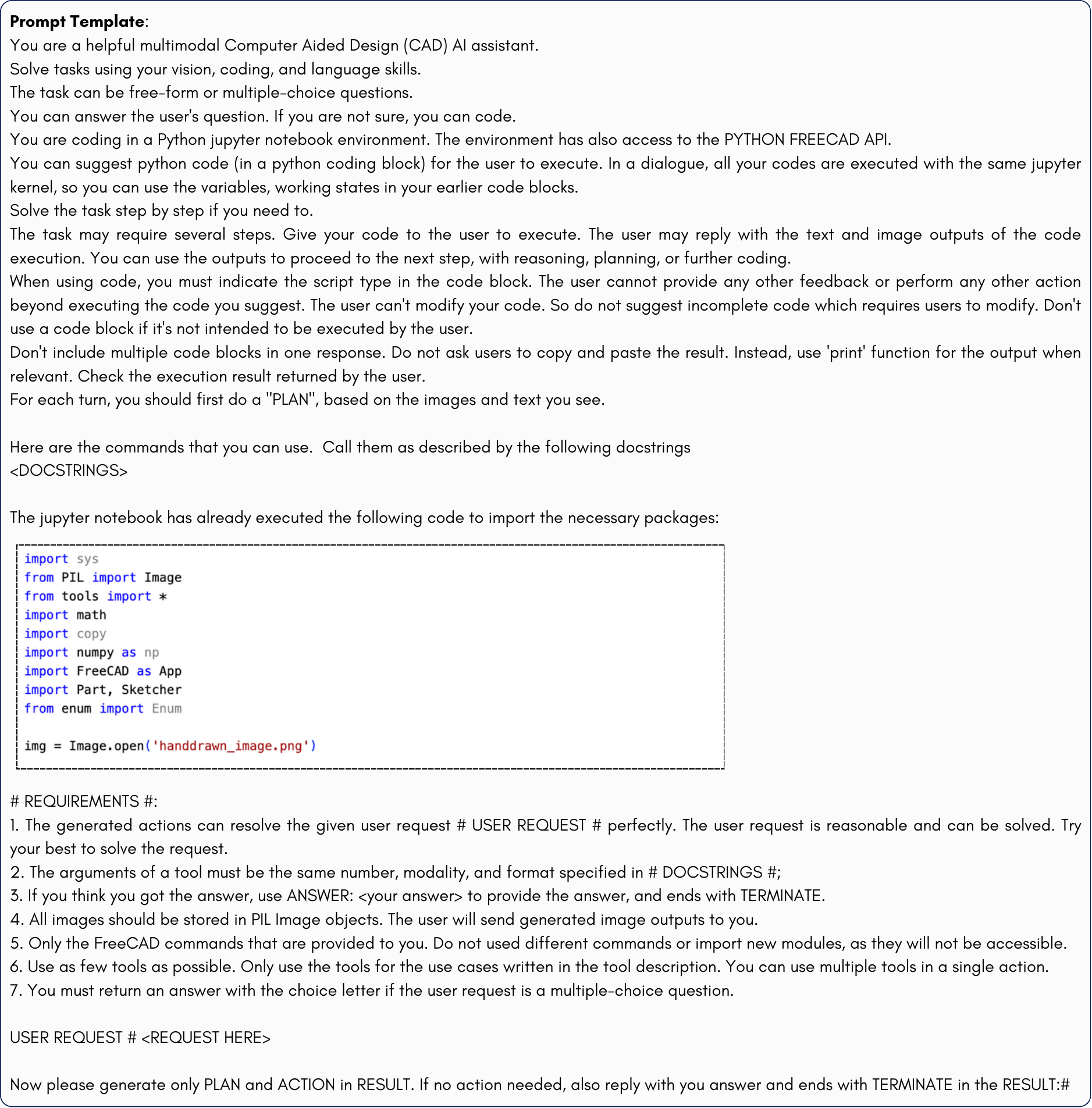}
    \caption{Prompt template for the \texttt{CAD-Assistant}. A detailed docstring disambiguating the use of the FreeCAD API and CAD-specific tools is provided as part of the prompt. The docstring is shown in section \ref{sec:docstring}.  On this example, the VLLM planner has a handdrawn sketch image preloaded. For different usecases, loading can be 3D scans or FreeCAD project files.}
    \label{fig:context}
\end{figure*}

In this work, we use a unified prompt template, similar to ~\cite{Hu2024VisualSS} for all CAD-specific problems. The prompt consists of three key components: (1) a general context, (2) a list of tools provided to the VLLM planner via docstrings, and (3) a multimodal user request. A summary of the FreeCAD API commands is provided in Table \ref{tab:freecad}, and the full set of docstrings supplied to the planner is presented in Section \ref{sec:docstring}. Note that as the set of considered API commands increases, the input context of the VLLM planner could increase. To address this, a preprocessing step could be implemented to dynamically select relevant docstrings before execution. The general context available to the VLLM planner is shown in Figure \ref{fig:context}.

\section{Beyond Simplified CAD Commands}

Extending the discussion of Sec. \ref{sec:beyond}, we provide an additional qualitative example of the proposed \texttt{CAD-Assistant}. Figure~\ref{fig:beyond_v2} shows the utilization of the CAD operation \textit{Fillet} by our method. It can be observed that \texttt{CAD-Assistant} computes the intersection of the lines to be able to perform the \textit{Fillet} operation on the corners by analyzing its docstring. Moreover, we find that VLLM planner performance might vary across CAD commands. This highlights the necessity of developing CAD-specific benchmarks tailored to CAD agents. Such benchmarks are crucial for gaining deeper insights into the capabilities and limitations of VLLM planners on generic CAD task solving.

\section{Qualitative Evaluation}

This supplementary material presents examples of complete agent trajectories for the CAD benchmarks used in this study. Detailed examples from the 2D and 3D subsets of SGPBench are provided in subsections \ref{sec:2Dres} and \ref{sec:3Dres}. Trajectories for the autoconstraining task are illustrated in subsection \ref{sec:autoconstrres}, while examples of hand-drawn parameterization are presented in subsection \ref{sec:paramres}.

\clearpage

\begin{table*}[t]
    \centering
    \setlength{\tabcolsep}{8pt}
    \begin{tabular}{l|lll}
        \toprule
         \textbf{Task} & \textbf{Avg Input Tokens} & \textbf{Avg Output Tokens} & \textbf{Avg Cost per User Request} \\
        \midrule
        CAD Question Answering & 11280 & 178 & \$0.0299\\
        Autoconstraining & 28422 & 852 & \$0.0795\\
        Handdrawn sketch parameterization & 31170 & 1081 & \$0.0887\\
        \bottomrule
    \end{tabular}
    \caption{Cost per user request for the \texttt{CAD-Assistant} utilizing GPT-4o as VLLM planner.}
\label{tab:costs}
\end{table*}

\begin{table*}[t]
    \centering
    \setlength{\tabcolsep}{3pt}
    \begin{tabular}{lp{0.4\textwidth}p{0.35\textwidth}}
        \toprule
         \textbf{FreeCAD Class} & \textbf{Class Methods} & \textbf{Class Attributes} \\
        \midrule

       \small{\texttt{Sketcher.Sketch}}  & \scriptsize{\code{\_\_init\_\_()}, \code{recompute()}, \code{delGeometries(indx)},
       \code{addConstraint(const)}, \code{addGeometry(geometry)}
       }
       &\scriptsize{\code{Name}, \code{Geometry},
       \code{Constraints},
       \code{State},
       \code{ConstraintCount},
       \code{GeometryCount},  \code{Placement}}
       \\
\midrule
         \small{\texttt{Sketcher.Constraints}} & \scriptsize{\code{\_\_init\_\_(constraintType, *args)}}  & \scriptsize{\code{Name}}\\
\midrule
         \small{\texttt{Part.Circle}}  & \scriptsize{\code{\_\_init\_\_(center, normal, radius)}} & \scriptsize{\code{Center}, \code{Radius}}\\
\midrule
         \small{\texttt{Part.Point}}  & \scriptsize{\code{\_\_init\_\_(point)}}& \scriptsize{\code{X}, \code{Y}, \code{Z}}\\
\midrule
          \small{\texttt{Part.ArcOfCircle}}  & \scriptsize{\code{\_\_init\_\_(circle, startParam, endParam)}}, \scriptsize{\code{\_\_init\_\_(startPoint, endPoint, midPoint)}}& \scriptsize{\code{Center}, \code{Radius}, \code{StartPoint}, \code{EndPoint},
         \code{FirstParameter}, \code{LastParameter}} \\
\midrule
         \small{\texttt{Part.LineSegment}}&\scriptsize{\code{\_\_init\_\_(startPoint, endPoint)}} & \scriptsize{\code{StartPoint}, \code{EndPoint}} \\
\midrule
         \small{\texttt{Part.Extrude}}&\scriptsize{\code{\_\_init\_\_()}} & \scriptsize{\code{Base},
         \code{DirMode},
         \code{LengthFwd},
         \code{LengthRev},
         \code{Solid},
         \code{Reversed},
         \code{Symmetric},
         \code{TaperAngle},\code{TaperAngleRev}} \\
\midrule
         \small{\texttt{Part.Solid}}&\scriptsize{\code{fuse(shape)}, \code{cut(shape)}, \code{common(shape)}} & \scriptsize{\code{TypeId}, \code{Volume}, \code{BoundBox}} \\
        \\[-10pt]

        \bottomrule
    \end{tabular}
    \caption{Summary of FreeCAD API classes, methods, and attributes utilized by the \texttt{CAD-Assistant} framework. The VLLM planner is supplied with docstrings that clarify their use, including detailed descriptions, function signatures and usage examples.}
\label{tab:freecad}
\end{table*}

\onecolumn
\subsection{More qualitative results on CAD question answering for the 2D Subset of SGPBench.}
\label{sec:2Dres}

\begin{figure*}[h]
    \centering
    \includegraphics[width=0.9\linewidth]{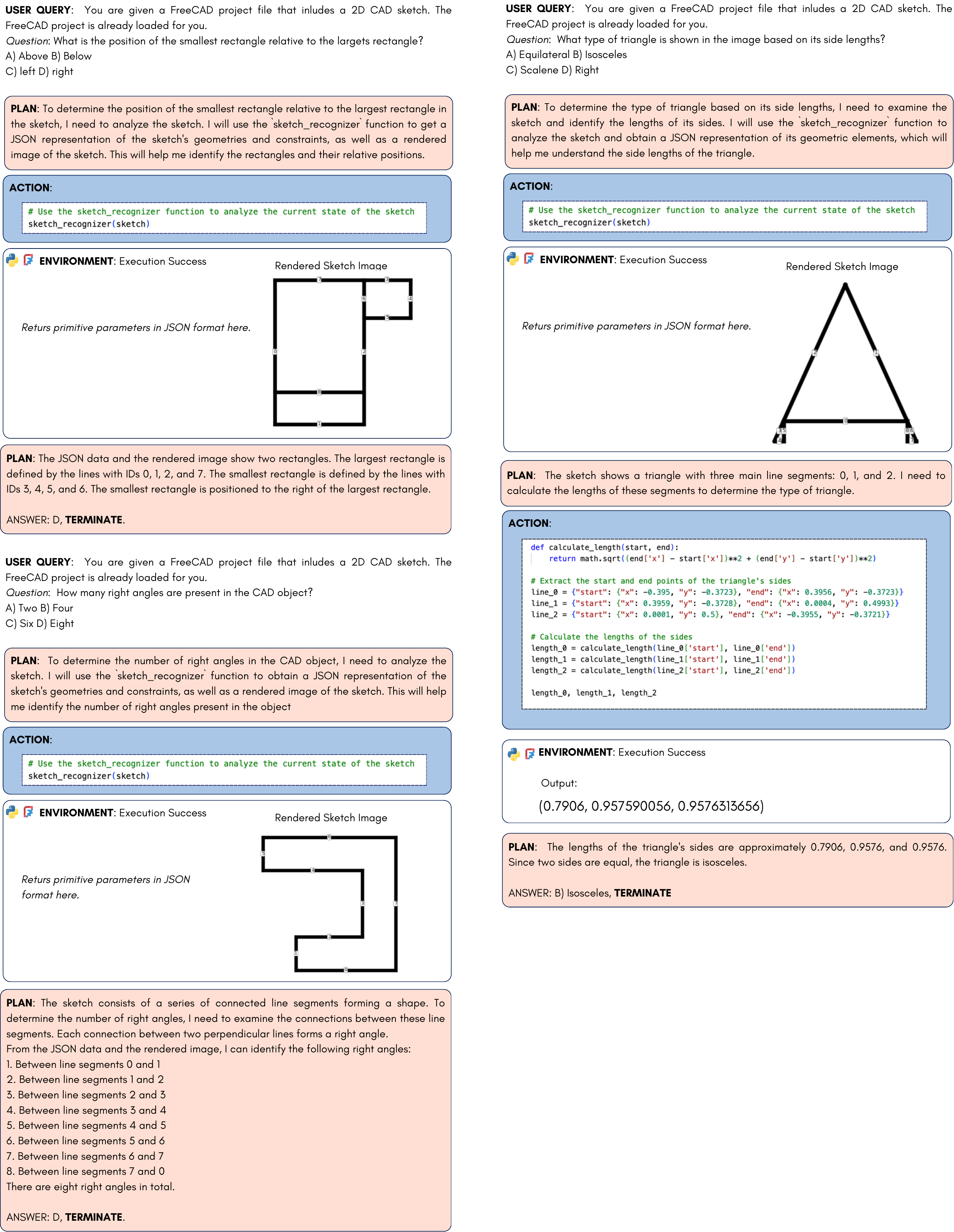}
    \caption{Complete agent trajectories of the \texttt{CAD-Assistant} for CAD Question Answering on the 2D subset of SGPBench.}
\end{figure*}

\clearpage

\subsection{More qualitative results on CAD question answering for the 3D Subset of SGPBench.}
\label{sec:3Dres}
\begin{figure*}[h]
    \centering
    \includegraphics[width=0.95\linewidth]{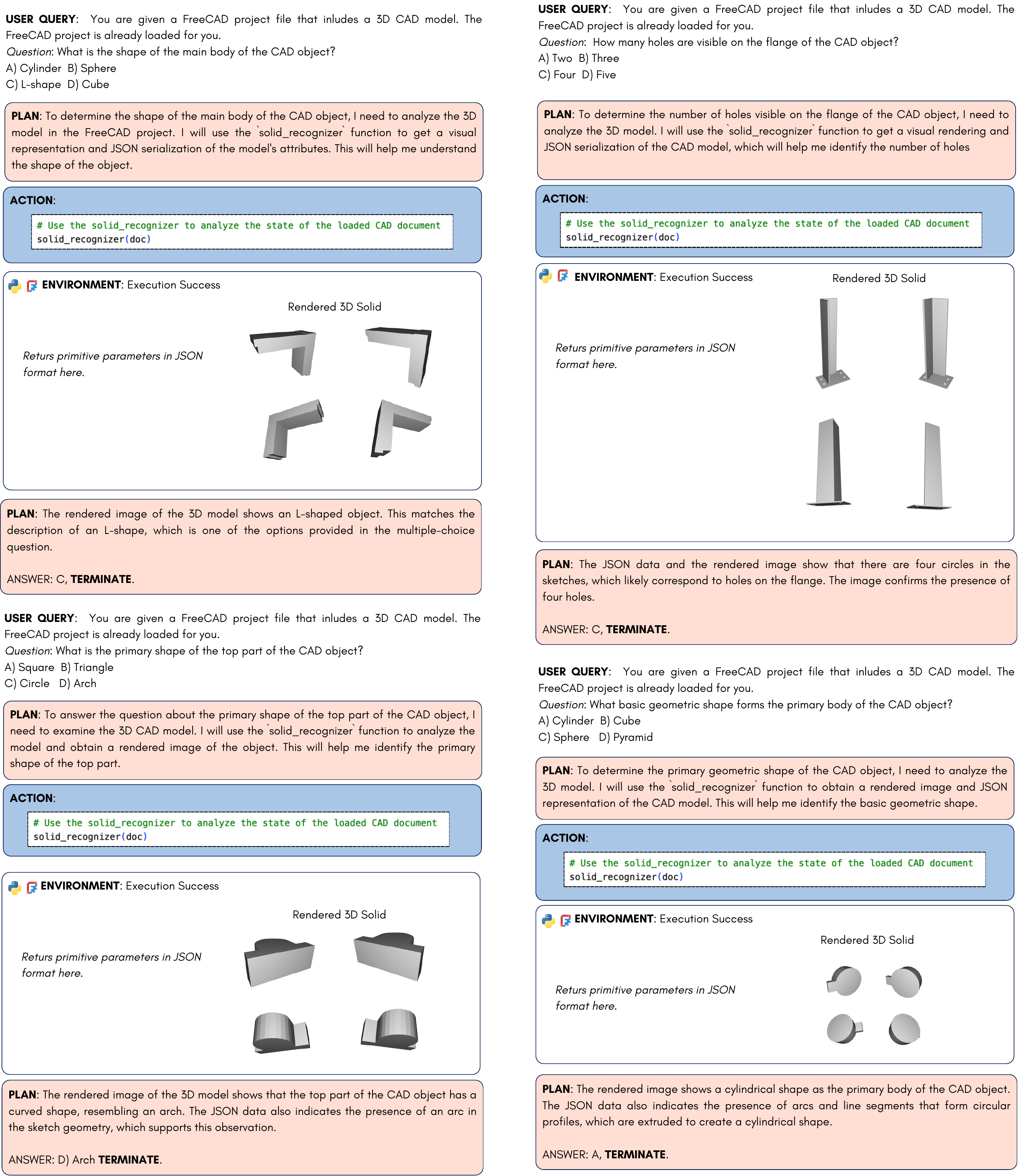}
    \caption{Complete agent trajectories of the \texttt{CAD-Assistant} for CAD Question Answering on the 3D subset of SGPBench.}
\end{figure*}
\clearpage

\subsection{More qualitative results on CAD sketch autoconstraining.}
\label{sec:autoconstrres}
\begin{figure*}[h]
    \centering
    \includegraphics[width=\linewidth]{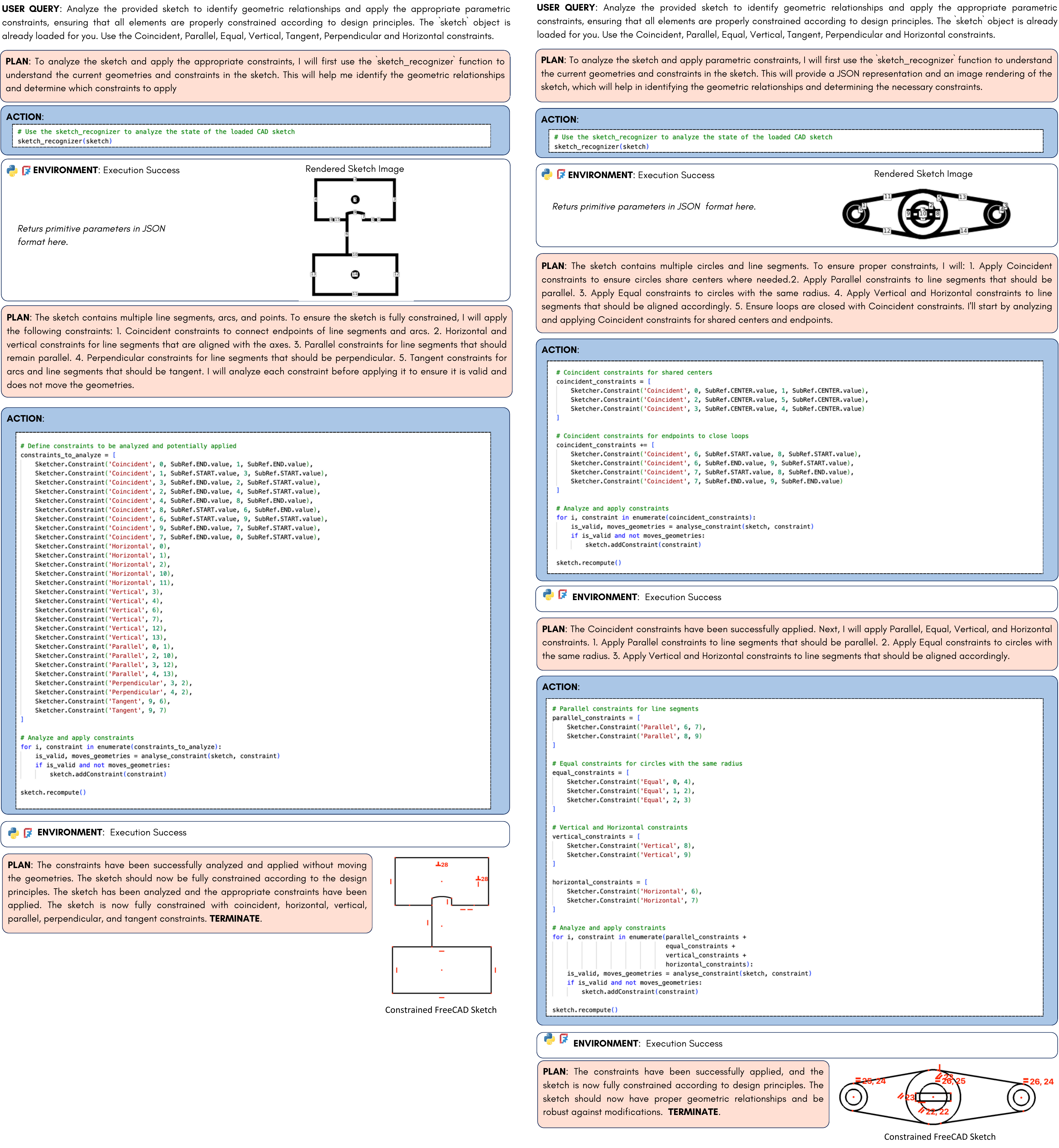}
    \caption{Complete agent trajectories of the \texttt{CAD-Assistant} for CAD sketch autoconstraining}
\end{figure*}
\clearpage

\subsection{More qualitative results on handdrawn CAD sketch parameterization.}
\label{sec:paramres}
\begin{figure*}[h]
    \centering
    \includegraphics[width=\linewidth]{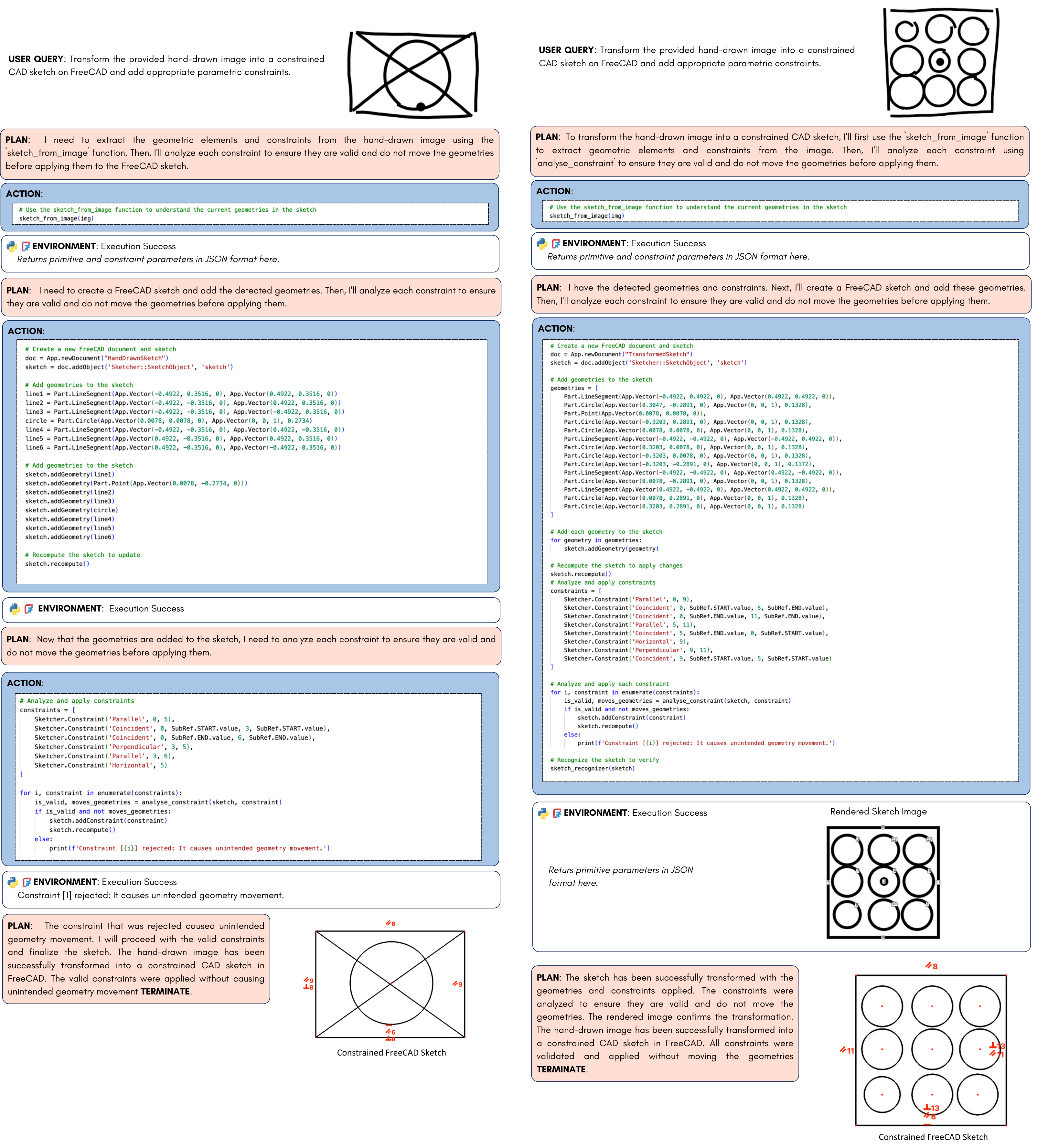}
    \caption{Complete agent trajectories of the \texttt{CAD-Assistant} for handdrawn CAD sketch parameterization.}
\end{figure*}
\clearpage

\onecolumn
\section{Docstrings}
\label{sec:docstring}
This section provides the complete docstring of the toolset available to the VLLM planner.
\begin{figure*}[h]
    \centering
    \includegraphics[width=0.93\linewidth]{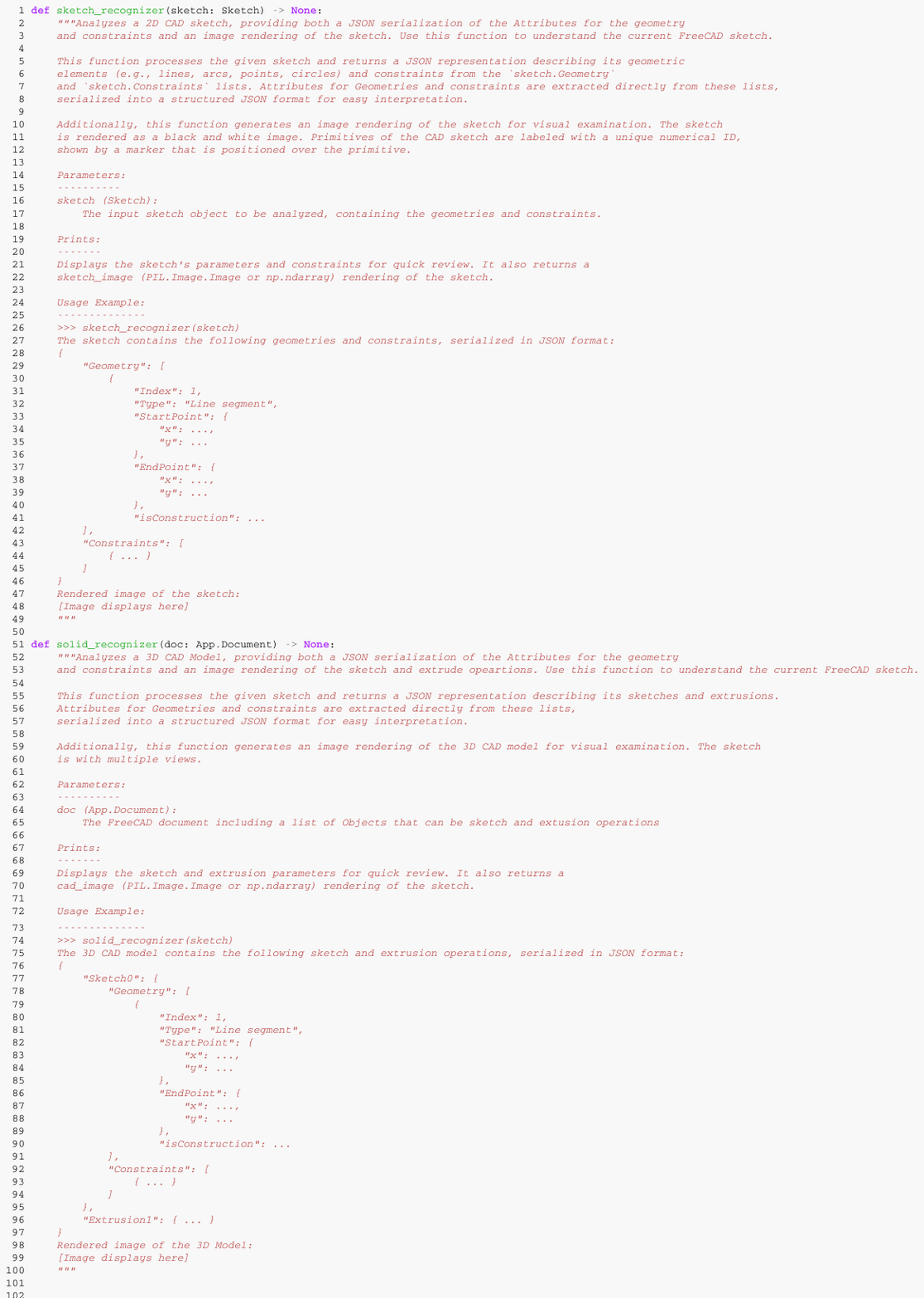}
\end{figure*}

\begin{figure*}[t]
    \centering
    \includegraphics[width=0.93\linewidth]{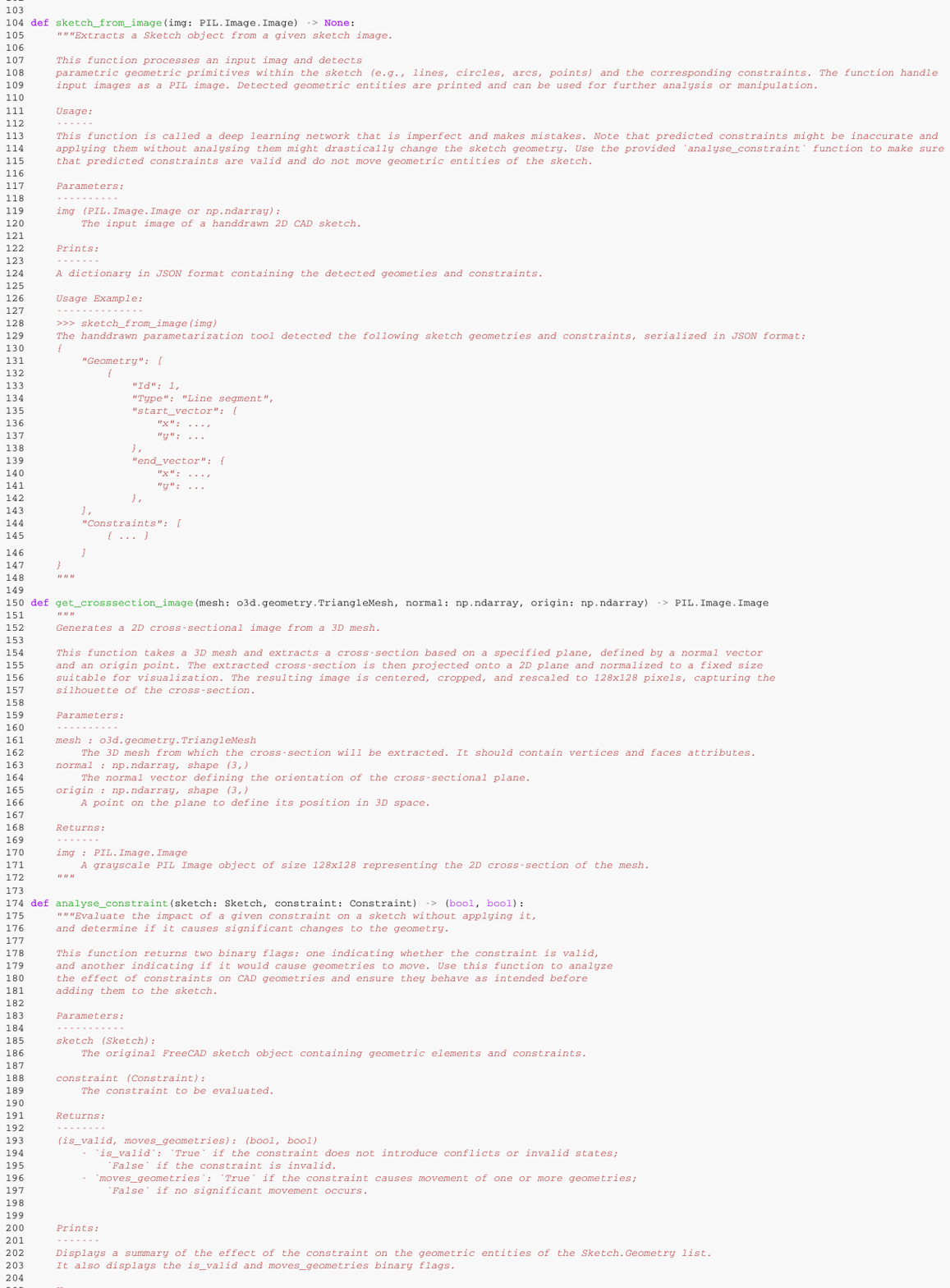}
\end{figure*}

\begin{figure*}[t]
    \centering
    \includegraphics[width=0.93\linewidth]{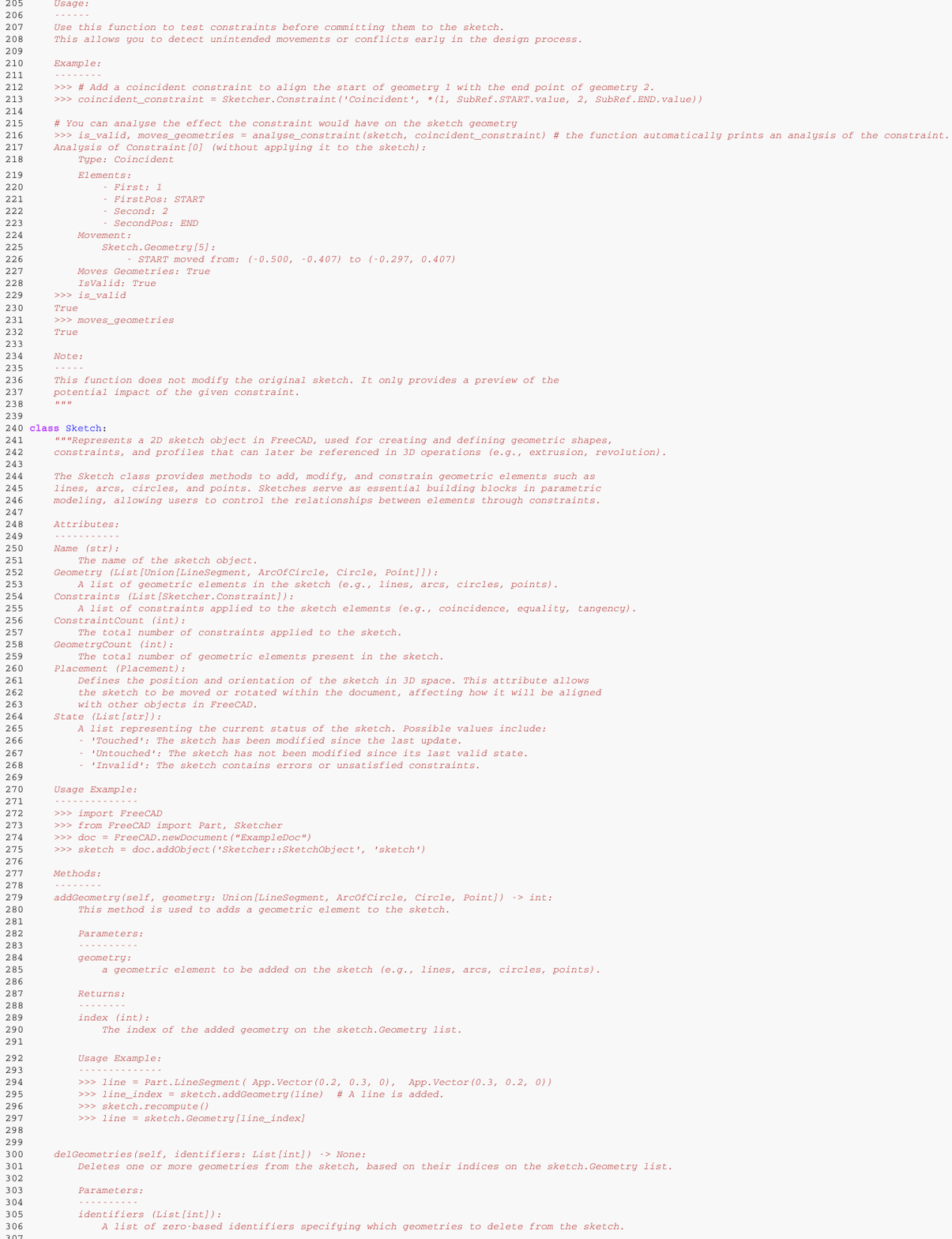}
\end{figure*}

\begin{figure*}[t]
    \centering
    \includegraphics[width=0.93\linewidth]{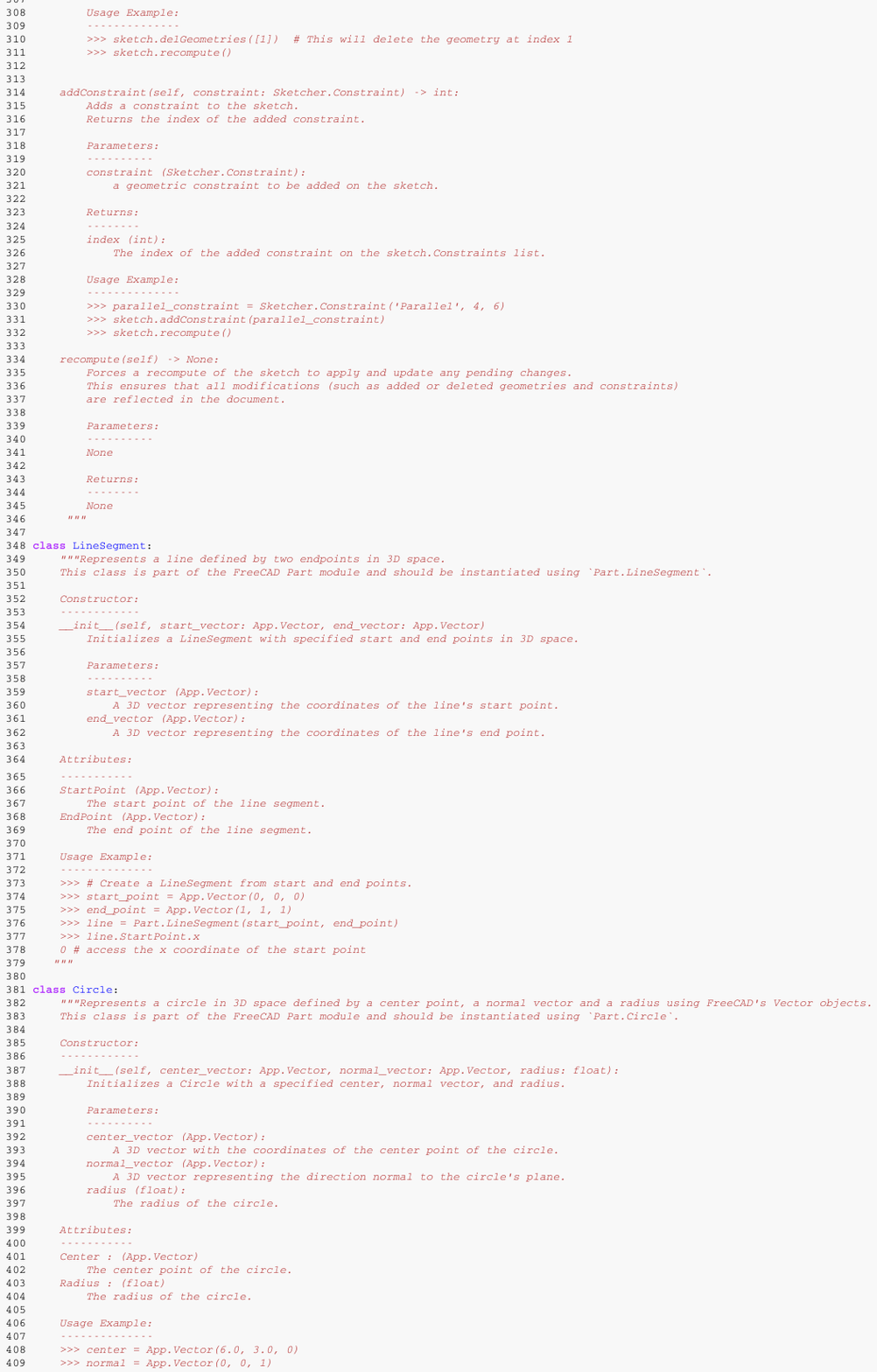}
\end{figure*}

\begin{figure*}[t]
    \centering
    \includegraphics[width=0.93\linewidth]{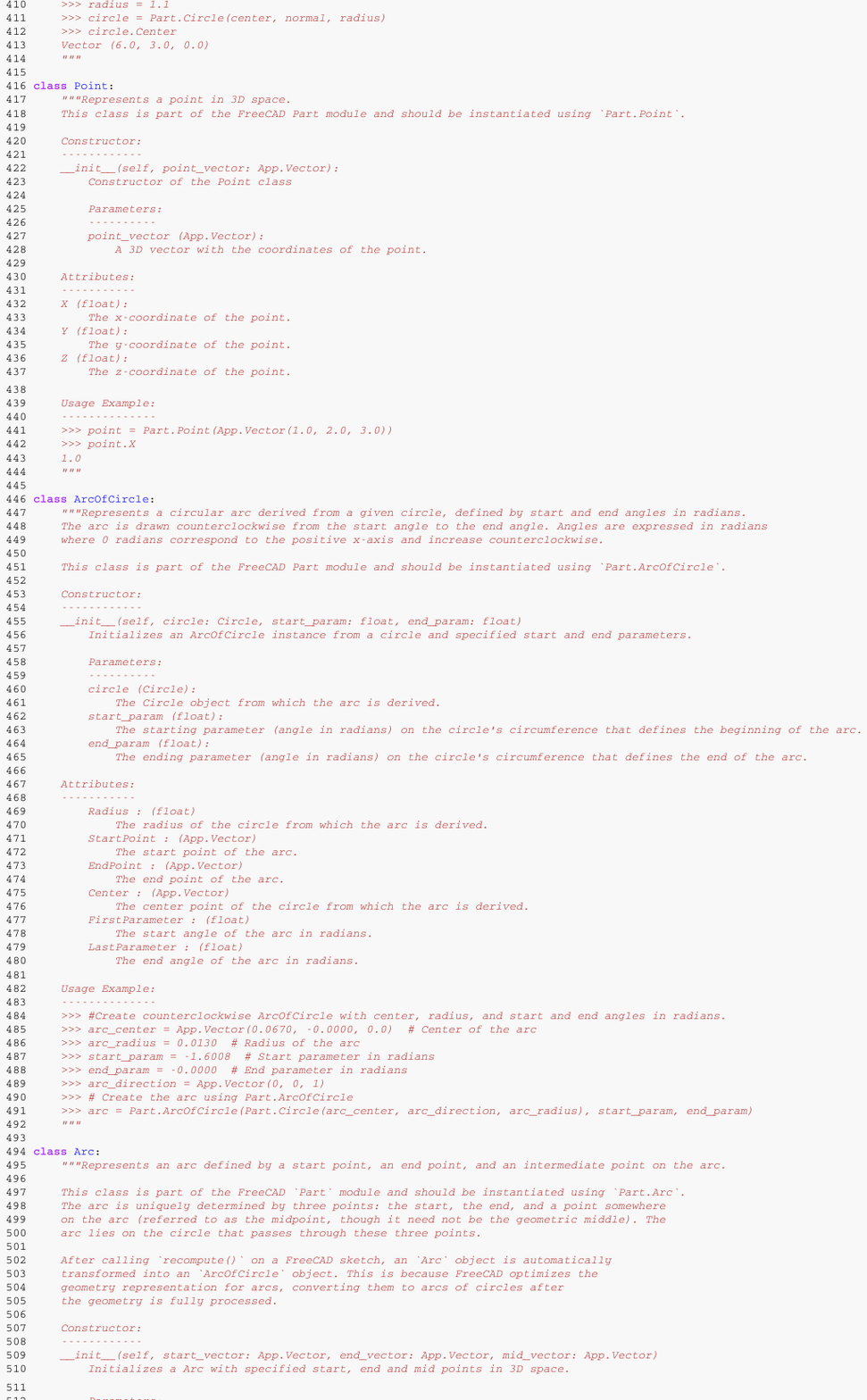}
\end{figure*}

\begin{figure*}[t]
    \centering
    \includegraphics[width=0.93\linewidth]{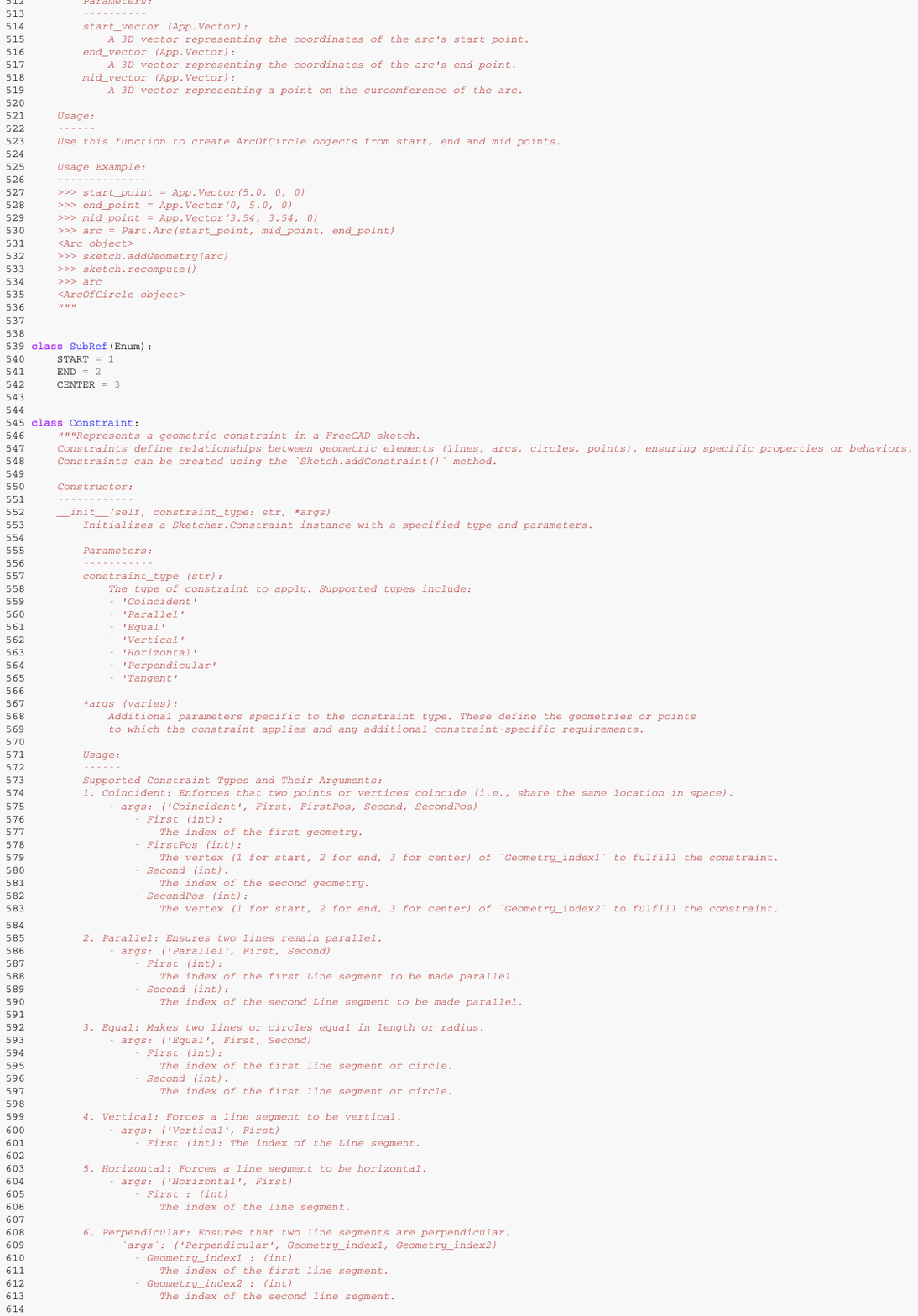}
\end{figure*}

\begin{figure*}[t]
    \centering
    \includegraphics[width=\linewidth]{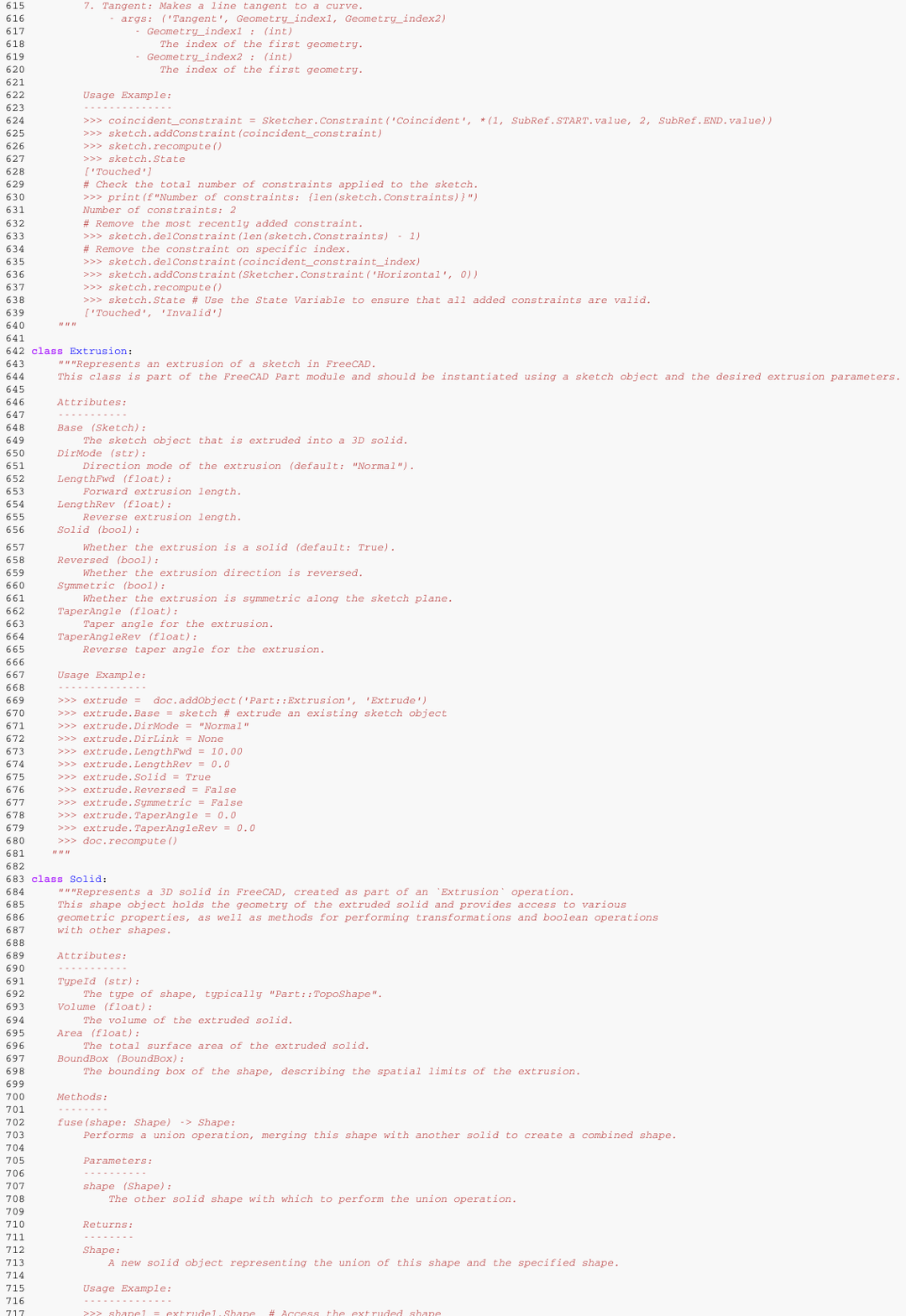}
\end{figure*}

\begin{figure*}[t]
    \centering
    \includegraphics[width=\linewidth]{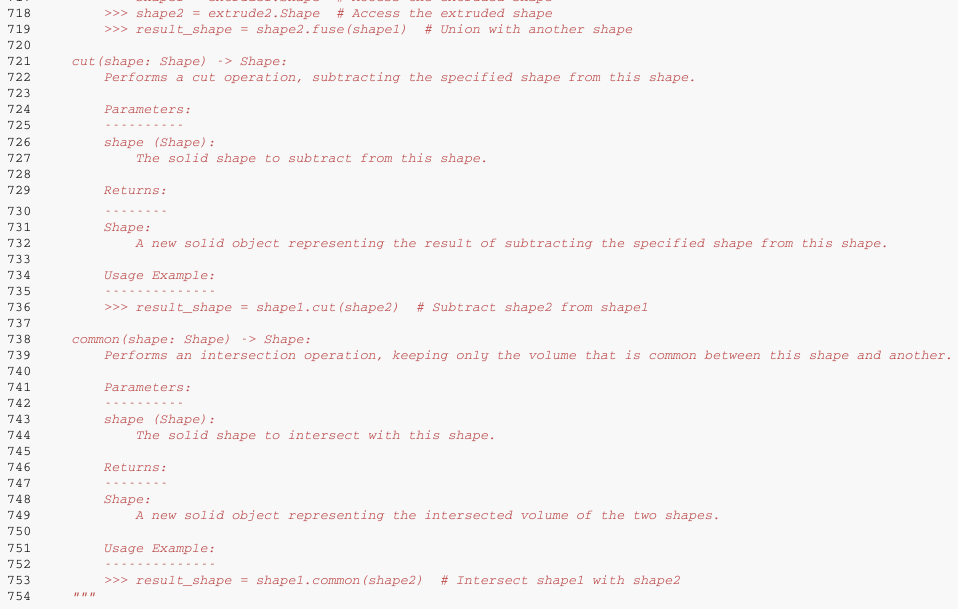}
\end{figure*}

\end{document}